\documentclass[journal]{IEEEtran}
\usepackage[utf8]{inputenc}
\usepackage{amsmath,graphicx}
\usepackage{nicefrac}
\usepackage{subcaption}
\usepackage{soul}
\usepackage{booktabs}
\usepackage{dblfloatfix} 
\usepackage{color}
\usepackage{setspace}
\usepackage[inline]{enumitem}
\usepackage{hyperref}
\usepackage{multirow}
\definecolor{darkgreen}{rgb}{0,0.6,0.2}

\usepackage{algorithm}
\usepackage{algorithmic}
\usepackage{array}

\usepackage{url}

\title{Unsupervised Image Regression for Heterogeneous Change Detection}

\author{Luigi~T.~Luppino,~
        Filippo~M.~Bianchi,~
        Gabriele~Moser,~
        and~Stian~N.~Anfinsen~
\thanks{L.T.~Luppino, F.M.~Bianchi and S.N.~Anfinsen are with the Machine Learning Group, Department of Physics and Technology, UiT The Arctic University of Norway, e-mail: luigi.t.luppino@uit.no.}
\thanks{G.~Moser is with DITEN Department, University of Genoa, Italy.}
}

\begin{document}

\maketitle

\begin{abstract}
Change detection in heterogeneous multitemporal satellite images is an emerging and challenging topic in remote sensing.
In particular, one of the main challenges is to tackle the problem in an unsupervised manner.
In this paper we propose an unsupervised framework for bitemporal heterogeneous change detection based on the comparison of affinity matrices and image regression.
First, our method quantifies the similarity of affinity matrices computed from co-located image patches in the two images.
This is done to automatically identify pixels that are likely to be unchanged.
With the identified pixels as pseudo-training data, we learn a transformation to map the first image to the domain of the other image, and vice versa.
Four regression methods are selected to carry out the transformation: Gaussian process regression, support vector regression, random forest regression, and a recently proposed kernel regression method called homogeneous pixel transformation.
To evaluate the potentials and limitations of our framework, and also the benefits and disadvantages of each regression method, we perform experiments on two real data sets.
The results indicate that the comparison of the affinity matrices can already be considered a change detection method by itself.
However, image regression is shown to improve the results obtained by the previous step alone and produces accurate change detection maps despite of the heterogeneity of the multitemporal input data.
Notably, the random forest regression approach excels by achieving similar accuracy as the other methods, but with a significantly lower computational cost and with fast and robust tuning of hyperparameters.
\end{abstract}

\begin{IEEEkeywords}
unsupervised change detection, multimodal image analysis, heterogeneous data, image regression, affinity matrix, random forest, Gaussian process, support vector machine, kernel smoothing
\end{IEEEkeywords}

\section{Introduction}

\IEEEPARstart{C}{hange} detection (CD) is a well known task in satellite remote sensing: the goal is to recognise changes in imagery acquired on the same location but at different times.
The applications range from damage assessment (bitemporal case) after natural disasters, such as floods, earthquakes, forest fires, and landslides, to long term trend monitoring (multitemporal case) of for example land cover and land use, urban development, glacier dynamics, deforestation, and desertification~\cite{lu2004change}.

Most of the past works on CD assume that the satellite images are homogeneous, i.e.\ the data were collected by the same kind of sensors, using the same configurations and modalities~\cite{zhang2016cd,zhao2017discriminative,gong2016coupled}.
These approaches may allow slight differences between the image domains at the two times, e.g.\ variations in light conditions or sensor geometry, but assume that it is possible to compensate for them by normalization, co-calibration or similar means~\cite{tuia2013graph}.
Even though there are techniques which mitigate the issues due to spatial misalignment~\cite{zhang2016cd,marcos2016geospatial,liu2016unsupervised}, co-registration is another fundamental assumption in CD: every pixel of the image at time one and its corresponding pixel of the image at time two are assumed to represent the exact same location on the Earth.

The methodological evolution has eventually turned the focus towards CD algorithms that allow the remote sensing community to fully exploit the large fleet of available sensors and constellations by considering joint use of heterogeneous satellite images.
This has led to the development of heterogeneous CD methods, which are those methods based on heterogeneous sources of data, also referred to as multisource, multimodal, multisensor, cross-sensor and information unbalanced data.
For example, a multispectral optical image combined with a polarimetric synthetic aperture radar (SAR) image represents the most common scenario \cite{zhao2017discriminative}.

As reviewed in~\cite{gong2016coupled}, there is not a unique way to group CD methods.
However, the distinction between techniques designed for homogeneous and heterogeneous data is clear.
The latter case invalidates the assumptions that the same physical quantities are measured, classes have always the same signatures, and data follow the same statistical behaviour~\cite{luppino2017clustering}.
Since traditional homogeneous CD techniques are unsuitable in this case, additional pre- or postprocessing steps are required~\cite{luppino2017clustering,liu2018change}.
To meet this challenge, several kinds of methodologies have been proposed.
These are related to topics such as domain adaptation, data transformation and transfer learning~\cite{marcos2016geospatial,liu2018change,liu2017change,tuia2016multi,khan2017forest}.
To summarise, the challenge is that we have bitemporal datasets which lie in different domains, and the task is to bring them into a common one where they can be compared.
To achieve this goal, supervised methods rely on labeled data, and this requirement is their strongest limitation: Such information is either the result of a meticulous and time-consuming manual selection, or the outcome of a potentially imprecise (if not unreliable) automatic delineation method.
On the other hand, the unsupervised framework is more challenging and prompts more complicated methodology due to the lower amount of prior information, but it does not require labelled data from unchanged areas as input.

In this work, we suggest a simple, yet effective methodology to perform unsupervised CD with data acquired under heterogeneous conditions or with heterogeneous sensors.
It is based on regression concepts and automatically generates the training dataset $\mathcal{T}$ for regression, which consists of $M$ corresponding pixel pairs from areas in the two images presumed not to be affected by changes.
According to~\cite{zhang2016cd,liu2018change}, the provision of such training data by manual selection would not be a strong requirement, although it prompts user interaction.
Nonetheless, besides being a costly and time-consuming operation, it also might lead to inaccurate information, especially when images are difficult to interpret~\cite{zhan2018iterative}.
Thus, we propose a completely unsupervised solution, which does not rely on any prior knowledge about the data.

Our approach is to first perform a preanalysis based on affinity matrices to provide an indication of where changes have happened.
Based on this, we select few data points from the estimated no-change areas.
This small pseudo-training set is used to learn a regression function which predicts how every pixel in one image domain would appear in the other image domain, and vice versa.
We will refer to this methodology as \textit{image regression}, a term which has on some occasions been used in the CD literature when translating between two more or less heterogeneous image domains~\cite{singh1989review,lunetta1999remote,mas1999monitoring}. In the machine learning literature, this is also known as image-to-image translation.
Once the predictions between the image domains have been computed, classic homogeneous CD methods can be applied to obtain change maps.
Concerning the regression function, we consider methods that are well-established, but new in this context: Gaussian process regression (GPR)~\cite{rasmussen2004gaussian}, support vector regression (SVR)~\cite{cortes1995support}, and random forest regression (RFR)~\cite{breiman2001random}.
We also compare with a kernel regression method, referred to as the homogeneous pixel transformation (HPT) by the authors in~\cite{liu2018change}, where it was used for the same purpose although in a supervised setting. This algorithm represents the state-of-the-art.

We test the consistency and the performance of the proposed regression methods, as well as their benefits and drawbacks, on two different datasets.
Our goal is to guide the user in choosing the most suitable approach according to requirements, whether the priority is best performance in change detection, easy tuning of hyperparameters, or short training and prediction time.

The remainder of this article is the following: Section \ref{sec:related} reviews the current literature on heterogeneous CD.
Section \ref{sec:method} introduces the reader to the proposed methodology, the notation, and the regression methods listed above.
Results on two datasets are presented in Section \ref{sec:results}.
Section \ref{sec:concl} concludes the paper.

\section{Related work}\label{sec:related}

To frame our work within the state-of-the-art, we try to classify methods from the literature according to how they tackle the heterogeneous CD problem, while limiting the scope of the review to the case of bitemporal data.
We group the methods into supervised and unsupervised and, for the latter, we further distinguish between traditional and deep learning approaches.

\subsection{Supervised methods}

The first attempts at heterogeneous CD were all supervised methods, such as the pioneering approach of Mercier et al.~\cite{mercier2008conditional}, who used quantile regression and copulas to build local models of dependence between unchanged areas in heterogeneous images.
The algorithm yields an estimate of the local statistics of the first image through the point of view of the second one, and compares these statistics via a Kullback-Leibler-type distance.

The copula-based approach is an example of a method which maps data from one input image domain to the other.
Another example is the more recent HPT method presented by Liu et al.~\cite{liu2018change}, which uses kernel regression on a sample of nearest neighbour pixels to set up mappings between the input domains.
The regression function is learned with training data labelled as unchanged.
The algorithm is explained in detail in the methodology section, since it is chosen as a reference algorithm representing the state-of-the-art.
The choice is justified by the fact that our work also belongs to the category of methods that map between the input data spaces.

An alternative methodology is to transform data from the heterogeneous input domains into a joint domain where CD can be performed by simple comparison.
The first of these approaches was proposed by Storvik et al.~\cite{storvik2009combination}, who obtained a joint distribution of the heterogeneous images by transforming their marginal densities in no-change areas into meta-Gaussian distributions, which provide simple and efficient models of multitemporal correlations.

Post-classification CD consists of selecting designated classifiers for the pre-event and the post-event data and mapping the data to a common label space. The classification maps are compared to identify the pixels which do not belong to the same class at both times. An example is found in~\cite{liu2014change}, where multidimensional evidential reasoning is defined and exploited to design a constrained model by which the probability of transitions between classes across the two images are computed.
Clearly, the performance of post-classification approaches depends highly on the choice and the design of the two classifiers, as well as on the quality and the size of the training set~\cite{zhao2017discriminative,zhan2018iterative}. It may therefore be applicable when targeting specific tasks, but is more questionable as a general approach.

A method exploiting a common manifold space is shown in~\cite{prendes2015new}. Here, the physical properties of the considered sensors and, especially, the associated measurement noise models and local joint distributions are used to define a no-change manifold. The latter is obtained by starting from a selection of patches that cover unchanged areas, from which the method learns to match the signatures of the involved classes in the two images. Finally, an ad hoc manifold distance is defined and the patches are projected into the manifold space, where those that lie too far from the no-change manifold are classified as changed areas.

In~\cite{tuia2016multi}, a semisupervised manifold alignment method based on kernels allows extraction of features to project the data from all the available sources into a joint latent space. This is achieved by forcing the local geometry of each manifold to remain unchanged. The method penalises projections that map samples of the same class far from each other and maximises a class dissimilarity term, which forces samples from different classes to be distant.

A kernel extension of canonical correlation analysis is proposed in~\cite{volpi2015spectral} to align data spaces. The idea is to first transform the data into a common Hilbert space, where the canonical correlation analysis can work in its original linear fashion. The parameter selection for the kernel transformation and the optimisation of the linear rotation-based canonical correlation analysis are automatic. However, the whole process relies on maximisation of the correlation between samples which are manually selected from unchanged areas.

\subsection{Unsupervised methods}

The main challenge in unsupervised heterogeneous CD lies in the identification of unchanged areas, which can be used to define mappings and projections that allow a comparison of the data from different image domains. The learning of these mappings will suffer if data samples from changed areas are included.

\subsubsection{Traditional methods}

Gong et al.\ came first when they presented an iterative coupled dictionary learning method in~\cite{gong2016coupled}. It learns two coupled dictionaries for the bitemporal images, that can be used to find sparse codes for co-located image patches. The codes consist of the weights used to reconstruct the patches from the dictionaries.
In the learning phase, random pairs of co-registered image patches are encoded with a single code, which jointly indexes both dictionaries. A first step in the iterations is to discard the coupled patches that produce the largest reconstruction error, since they are more likely to cover changed areas. The remaining patches are used to iteratively update the dictionaries, thereby minimising the reconstruction error. In this way, the two dictionaries are brought to produce similar sparse codes for patches covering unchanged areas. During the test phase, the co-located patches are encoded separately and assigned individual codes. If these are dissimilar, the area covered is classified as changed.

Another significant contribution is the energy-based model encoding nonlocal pairwise pixel interactions, as proposed by Touati and Mignotte~\cite{touati2018energy}. This approach estimates a robust similarity feature map, resulting from the optimisation of a global cost function which enforces consistency with local pairwise pixel similarities (or dissimilarities) and their bitemporal evolution. The similarity map is segmented by multiple automatic threshold algorithms before a majority vote brings about the final ensemble result.

In~\cite{touati2018change}, the features of a pixel are substituted with the statistics of the histogram and the gradient magnitude histogram of its neighbourhood.
Benefits include the addition of spatial information in the analysis and the attenuation of noise, especially the multiplicative noise typical of SAR images.
The two high-dimensional feature spaces which the data are projected into are then reduced to one dimension by a fast implementation of the \textit{multidimensional scaling} algorithm.
The latter insures that pixels being different (similar) in the two high-dimensional spaces remain different (similar) to a certain proportion in the two one-dimensional spaces.
However, it does not ensure that changed (unchanged) areas are represented on the two images with different (similar) grey-level intensities.
To correct this issue, histogram matching is applied.
Finally, a pixel-wise difference image between the two one-dimensional representations is computed and then three different automatic thresholding algorithms are combined to obtain the final segmentation into the change and no-change classes.

\subsubsection{Deep learning methods}

In addition to these notable examples, the exponentially increasing interest in deep learning has also led to the development of novel architectures, both for homogeneous~\cite{khan2017forest,lyu2016learning} and heterogeneous CD~\cite{zhang2016cd,zhao2017discriminative,su2017deep}. Most of these methods are examples of feature learning, since they exploit the capability of e.g. convolutional neural networks and especially stacked denoising autoencoders (SDAEs) to infer spatial information from the data and learn new representations.

In~\cite{zhao2017discriminative}, discriminative features are learned from the heterogeneous images by introducing approximately symmetrical convolutional coupling networks.
These assure that pixels from unchanged (changed) areas are mapped to similar (different) values in code space by minimising the difference between codes of unchanged pixels. This is obtained with a loss function which adds up pixel-wise differences weighted by the probability that a pixel is unchanged. In a two-step iteration, the coupling networks are first trained according to the pixel probabilities, before keeping them fixed and updating the probabilities based on how different the two computed codes are.
At the end of the procedure, the changed areas are highlighted by thresholding.

In a very similar fashion, \textit{Zhang et al.\ }proposed in~\cite{zhan2018iterative} to learn new representative features for the two images by the use of two distinct SDAEs.
Again, probability maps are initialised randomly and the training alternates between two phases: updating the parameters of the network according to the two maps, and updating the maps according to the output of the network. Instead of producing a binary change map, this method introduces a hierarchical clustering strategy which highlights different types of change as separate clusters.

In~\cite{su2017deep}, change vector analysis is carried out to distinguish unchanged areas, positive changes and negative changes, as defined in~\cite{bovolo2007theoretical}.
SDAEs are exploited to extract relevant features and transfer the data into a code space.
A clustering of code differences obtained from co-located patches in the two images allows a preliminary distinction between samples from the three categories. Three distinct stacked mapping networks are then trained to transform the code from image one into three plausible codes from image two according to the categorisation.
The three potential codes are compared to the original code from image two and the pixel is assigned to the class producing the smallest difference.

In~\cite{zhan2018log}, a log-transformation of SAR data allows such patches to be stacked together with corresponding patches of optical data.
A SDAE is then used to extract one feature map for each modality.
The feature maps are clustered separately and a difference image is obtained by comparing the results.
Further, the clustering difference image is segmented into pixels certain to belong to changed areas, pixels certain to belong to unchanged areas, and uncertain pixels.
Finally, the features of all the pixels belonging to the certain clusters are used to train a neural network which is then able to map the uncertain pixels to the right clusters, yielding to the final binary change map.

Although it is not a completely unsupervised method, it is worth to mention the work presented in~\cite{zhang2016cd}.
After a preliminary stage to improve the co-registration of the two images, post-classification CD yields a coarse CD map. 
This map is then used to select the parts of the images least likely to be affected by changes, which is needed to train a deep mapping network.
The latter is able to bridge the two different representations and to highlight changes, which are finally extracted via segmentation.


\section{Methodology}\label{sec:method}

In most of the unsupervised algorithms reviewed in Sec.\ \ref{sec:related}, a key step is to find a coarse CD map.
This allows either to select areas which are least likely to represent changes, so they can be used to learn how to map unchanged areas across the two domains, or to start from a reasonable point which can be iteratively improved.
In particular, unsupervised methods often rely on first initializing randomly the change map, and then improving it iteratively by enforcing that pixels which are least (most) likely to cover changed areas must have a low (high) intensity in such a heat-map.
Instead, we present in the following subsection a procedure which yields the same preliminary results in an automatic manner.

\subsection{Comparison of Affinity Matrices}\label{subsec:affm}

We follow the notation adopted in~\cite{liu2018change}: Two images that represent the same region are acquired by sensors $\mathcal{X}$ and $\mathcal{Y}$ at different times, $t_1$ and $t_2$.
The respective images are denoted as $\boldsymbol{X} \in \mathbf{R}^{n_1 \times n_2 \times P}$ and $\boldsymbol{Y} \in \mathbf{R}^{n_1 \times n_2 \times Q}$, where $n_1$ and $n_2$ are the common height and the width of the two images having $P$ and $Q$ channels, respectively. The common dimensions are obtained through resampling and co-registration.
We further assume that a limited part of the image has changed between time $t_1$ and $t_2$.
To learn the required data transformations, we seek a training set $\mathcal{T}=\{\left(\boldsymbol{x}_m,\boldsymbol{y}_m\right)\}_{m=1}^M$, selected from the set of all the $N=n_1 \cdot n_2$ pixels. Specifically, $\boldsymbol{x}_m=\left(x_{m,1},\dots,x_{m,P}\right)$ and $\boldsymbol{y}_m=\left(y_{m,1},\dots,y_{m,Q}\right)$ are the feature vectors of the $m$th training sample in $\boldsymbol{X}$ and $\boldsymbol{Y}$, respectively, and are composed of the pixel intensities in the corresponding $P$ and $Q$ image bands, respectively.

Now consider a pair of corresponding patches $p^{\boldsymbol{X}}$ and $p^{\boldsymbol{Y}}$ extracted over the same area $p$ from $\boldsymbol{X}$ and $\boldsymbol{Y}$, respectively.
The patches cover a $k \times k$ window, whose pixels $i=1,\dots,k^2$ can be indicated as $p^{\boldsymbol{l}}_{i}$, where the modality index $\boldsymbol{l}$ indicates either $\boldsymbol{X}$ or $\boldsymbol{Y}$.
For each patch $p$ and modality $\boldsymbol{l}$, the distance $d\left(p^{\boldsymbol{l}}_{i},p^{\boldsymbol{l}}_{j}\right)$ is computed between all pixel pairs $(i, j)$ in the patch.
There are several possible choices for the distance measure~\cite{luppino2017clustering}, whose appropriateness depends on the underlying data distribution.
We selected the computationally efficient Euclidean distance, which is suitable for data whose (multidimensional) distribution is nearly Gaussian.
This assumption is valid for imagery acquired by optical sensors~\cite{bovolo2007theoretical,bovolo2015time}.
SAR intensity data can also be brought to near-Gaussianity by a logarithmic transformation, as can other sensor data through the right choice of transformation~\cite{luppino2017clustering,zhan2018log}.

Once the distances are computed for all pixel pairs, these can be converted to corresponding affinities
\begin{equation}
A^{\boldsymbol{l}}_{i,j} =  \exp\left\{-\frac{d\left(p^{\boldsymbol{l}}_{i},p^{\boldsymbol{l}}_{j}\right)^2}{h^2}\right\}\,,
\label{eq:affm}
\end{equation}
that are the entries of the affinity matrix $A^{\boldsymbol{l}} \in \mathbf{R}^{k^2 \times k^2}$ for the given patch and modality.
The kernel width $h$ should be determined automatically.
Our choice is to compute the distance to the $7^{th}$ nearest neighbour for all data points in $p^{\boldsymbol{l}}$, and to set $h$ equal to the mean of these distances.
This heuristic captures the structure within the patch and it is robust with respect to outliers, because the values $A^{\boldsymbol{l}}_{i,j}$ of the closest neighbourhood are kept within a reasonable interval, whereas the rest gradually decay to $0$~\cite{myhre2012mixture}.
Other more common approaches have been tested, for example Silverman's rule of thumb~\cite{wand1995kernel}, but they did not prove as reliable as the selected one.

The two affinity matrices $A^{\boldsymbol{X}}$ and $A^{\boldsymbol{Y}}$ can be interpreted as containing the edges of two fully connected graphs with each pixel in the patch as a vertex.
They therefore hold rich information about the spatial structure and interrelations between pixels in the patch.
The core idea of this procedure is that if changes occur within the patch area, then the graph structure will change, including all the relations for the changed pixels.
Intuitively, the more changes occur in the patch, the more the two affinity matrices will diverge.
Hence, we need to quantify how dissimilar the affinity matrices of the different modalities are.

To quantify the distance between affinity matrices, we first obtain the element-wise difference $A^{\boldsymbol{X}} - A^{\boldsymbol{Y}}$. Then its matrix norm (also called Frobenius norm~\cite{horn1990matrix}) $f = \left\|A^{\boldsymbol{X}} - A^{\boldsymbol{Y}}\right\|$ is computed and assigned to all the pixels in the patch.
All the operations described above are performed on all the possible overlapping patches of the two images.
Therefore, for each pixel $n=1,\dots,N$ in the final CD map there is a set $F_n$ of Frobenius norms corresponding to the patches covering that pixel, and the final outcome for this specific pixel is the average of this set, which is stored as a possibility\footnote{The range of $0\leq P_c(n)\leq 1$ suggests it is a probability, but this interpretation is strictly not valid, so we call it a possibility.} that pixel $n$ is changed: $P_c(n)=\overline{F_n}$.

Algorithm \ref{alg:aff} summarises the whole procedure:
\begin{algorithm}
\begin{spacing}{1.2}
\caption{Possibilities of change for each pixel}\label{alg:aff}
\begin{algorithmic}
\FORALL{$p$}
\STATE Compute distances between all pixel pairs in $p^{\boldsymbol{X}}$
\STATE Compute distances between all pixel pairs in $p^{\boldsymbol{Y}}$
\STATE Determine $h^{\boldsymbol{X}}_p$ and $h^{\boldsymbol{Y}}_p$
\STATE Compute $A^{\boldsymbol{X}}$ and $A^{\boldsymbol{Y}}$
\STATE Compute $f = \left\|A^{\boldsymbol{X}} - A^{\boldsymbol{Y}}\right\|$ and add to $F_i$ for $\forall i \in p$
\ENDFOR
\FORALL{$n=1,\ldots,N$} 
\STATE Compute the mean over $F_n$ and assign to $P_c(n)$
\ENDFOR
\end{algorithmic}
\end{spacing}
\end{algorithm}

At this point, the $M$ pixels presenting the lowest $P_c(n)$ are selected as the training set $\mathcal{T}$.
Nevertheless, the fact that the $M$ pixels in $\mathcal{T}$ are the ones covered, on average, by the most similar affinity matrices across the two domain, may not mean that $\mathcal{T}$ is representative for the whole dataset and all surface types.
One of the physical elements of the image could be predominant in $\mathcal{T}$ (e.g., water or grass or bare soil) or some of these typologies may not be included in $\mathcal{T}$.
Intuitively, this may lead to an erroneous transformation of the pixels that are not represented in the training set, which will cause false positives in the change map.
As an unsupervised sanity check, we evaluate the Hellinger distance measure between the normalised histograms $\mathcal{H}_{\boldsymbol{l}}$ and $\mathcal{H}_{\boldsymbol{l} \cap \mathcal{T}}$ of $\boldsymbol{l}$ and $\boldsymbol{l} \cap \mathcal{T}$, for $\boldsymbol{l} = \boldsymbol{X},\boldsymbol{Y}$.
That is, we compute the distance between the histogram of one whole image domain and the training samples selected from the same image domain.
The Hellinger distance between two discrete one-dimensional histograms is defined as~\cite{cha2007comprehensive}:
\begin{equation}
    d_H=\sqrt{1-\sum_{i=1}^{N_{\textrm bins}}\sqrt{\mathcal{H}_{\boldsymbol{l}}(i)\cdot \mathcal{H}_{\boldsymbol{l} \cap \mathcal{T}}(i)}}\,,
\end{equation}
where the histograms are divided into $N_{\textrm bins}$ bins. The sum over $i$ is known as the Bhattacharyya coefficient.
To consider all the channels $C$ involved into a single distance, we modify the distance as follows:
\begin{equation}
    d_H=\sqrt{1-\frac{1}{C}\sum_{j=1}^{C}\sum_{i=1}^{N_{\textrm bins}}\sqrt{\mathcal{H}_{\boldsymbol{l},j}(i)\cdot \mathcal{H}_{\boldsymbol{l} \cap \mathcal{T},j}(i)}}.
    \label{eq:hell_mod}
\end{equation}
This diagnostic step checks that the training set $\mathcal{T}$ is representative, but it does not offer a solution in case it is not and the output of Eq.\ \ref{eq:hell_mod} is large.
A possible alternative sampling strategy could first divide the data across the entire feature space into clusters, and then select the $2\%$ of the pixels belonging to each cluster and presenting the lowest $P_c$.
Nevertheless, this suggested approach is not going to be tested in Sec.\ \ref{sec:results}, since it is out of the scope of this work.

We also note that the whole procedure summarised in Alg.\ \ref{alg:aff} needs a computation time which grows with the square of the patch size $k$.
A possible solution to mitigate such a problem is to reduce the total number of patches.
Instead of having a sliding window which covers every possible patch $p$ in the image, one can use a step size $\Delta>1$, thereby decreasing the number of patches by a factor $\Delta^2$.

The proposed method for measuring the structural similarity of multimodal image patches is closely related to the local self-similarity (LSS) approach introduced by Schechtman and Irani in~\cite{shechtman2007matching}. They compute affinities between patches instead of pixels, and compare a patch centered at the pixel-under-study with all possible equal-size patches in a wider search area. The resulting affinity surface is resampled on a log-polar grid, before the affinity samples in all grid cells are stacked into a feature vector representing the LSS of the given pixel. The LSS approach was extended to a global self-affinity (GSS) approach by Deselaers and Ferrari~\cite{deselaers2010global}, and both methods have been applied to various image matching tasks. This includes the application to co-registation of multimodal earth observation images by Sedaghat and Ebadi, who extended the LSS into a new image matching descriptor~\cite{sedaghat2015distinctive}, and Marcos et al., who constructed another image descriptor drawing upon both LSS and GSS~\cite{marcos2016geospatial}. Our proposed method can be seen as a special case of the patch-based LSS and GSS approaches, only with unit patch size. Another difference is that we compute a Frobenius norm that acts as a summary statistic for the patch-level difference between $p^{\boldsymbol{X}}$ and $p^{\boldsymbol{Y}}$, whereas the LSS and GSS feature vectors maintain a pixel resolution.

Noteworthy, the use of patch affinities may introduce more robustness towards image misregistration. The log-polar grid adds further resilience towards affine transformations and nonlinear warping effects that may occur in map projection and geocoding processes. Such measures could potentially improve the proposed method, but we uphold our initial assumption of perfect co-registration to focus on the change detection task, and argue that our method should in itself be reasonably robust to small registration errors as long as it is applied to medium resolution images and aimed at detecting changes in natural terrain that extend well beyond pixel-level.

\subsection{Image regression}

The training data $\mathcal{T}$ allows us to learn a regression function $f^{(1)}$ such that
 \begin{equation}
 \boldsymbol{\hat{y}}_m = f^{(1)}\left(\boldsymbol{x}_m\right) = \boldsymbol{y}_m + \epsilon^{(1)}_m, \quad \ m=1,\dots,M
 \end{equation}
where $\boldsymbol{\hat{y}}_m$ is the predicted value, $\boldsymbol{x}_m$ is the regressor, and $\epsilon^{(1)}_m$ is the residual.
We then train the reverse regression equation
 \begin{equation}
 \boldsymbol{\hat{x}}_m = f^{(2)}\left(\boldsymbol{y}_m\right) = \boldsymbol{x}_m + \epsilon^{(2)}_m, \quad \ m=1,\dots,M
 \end{equation}
in which $\boldsymbol{\hat{x}}_m$ is predicted starting from the regressor $\boldsymbol{y}_m$.
With these two functions it is possible to predict $\boldsymbol{\hat{Y}}$, i.e.\ the image which would have been obtained if sensor $\mathcal{Y}$ had observed the reality at time one, and $\boldsymbol{\hat{X}}$, the image of the reality at time two which would have been acquired by sensor $\mathcal{X}$.
Once the two predictions are computed, conventional change metrics such as image differences or ratios can be applied to highlight the differences between the original images and the corresponding predicted ones.
There is a plethora of effective homogeneous CD techniques which could be applied at this stage.
However, the main goals of this work are first to prove the reliability of what we will refer to as the self-supervised training set selection, and second to compare the image regression methods applied to obtain the predicted images. We now outline how we perform the hypothesis test to produce a final change map.

\begin{figure}[ht!]
\centering
\includegraphics[width=\columnwidth]{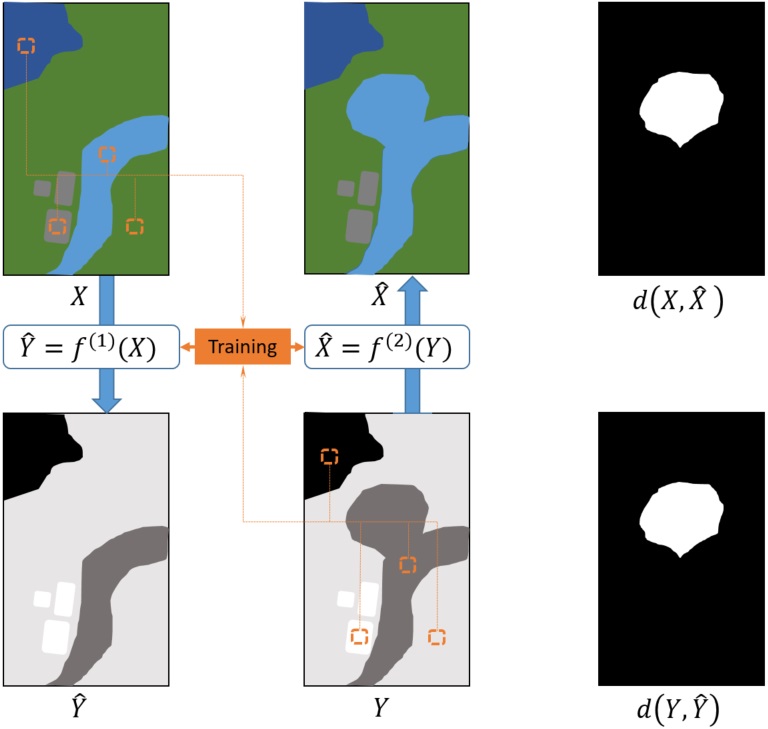}
\caption{Image regression: the two functions $f^{(1)}$ and $f^{(2)}$ are trained starting from the same data points, two predicted images are obtained, and finally two difference images are achieved.}
\label{fig:method}
\end{figure}

The two-way regression, consisting of the mappings $\boldsymbol{X}\to\hat{\boldsymbol{Y}}$ and $\boldsymbol{Y}\to\hat{\boldsymbol{X}}$, can be utilised in an ensemble approach where two weaker results are combined to obtain one stronger and more reliable outcome.
Let the distance images be defined as $\boldsymbol{D}_x=d(\boldsymbol{X},\hat{\boldsymbol{X}})$ and $\boldsymbol{D}_y=d(\boldsymbol{Y},\hat{\boldsymbol{Y}})$, where $d(\cdot,\cdot)$ is some distance measure $$d(\cdot,\cdot):\mathbf{R}^{n_1 \times n_2 \times C} \times \mathbf{R}^{n_1 \times n_2 \times C} \longrightarrow \mathbf{R}^{n_1 \times n_2}\,,$$ i.e.\ a pixel-wise distance between two images of size $n_1\! \times\! n_2$ and with $C$ channels ($C$ being either $P$ or $Q$).
If the distance images $\boldsymbol{D}_x$ and $\boldsymbol{D}_y$ are normalised and combined, distances that are consistently high in both images will indicate high probability of change, whereas false alarms due to a spurious high value in one of the distances will be suppressed.
We choose to combine the distances by a simple average: $\boldsymbol{D}=(\boldsymbol{D}_x+\boldsymbol{D}_y)/2$.
Before normalising $\boldsymbol{D}_x$ and $\boldsymbol{D}_y$, it is reasonable to clip the distances beyond some standard deviations of the mean value (e.g., $\boldsymbol{D}_x > \overline{\boldsymbol{D}_x}+4\sigma_{\boldsymbol{D}_x}$), so that outliers do not compromise such a step.
Fig.\ \ref{fig:method} illustrates the methodology.

At this stage, noise filtering may be applied.
In this work we choose to filter the difference image $\boldsymbol{D}$ by use of the method proposed in~\cite{krahenbuhl2011efficient} which uses spatial context to regularise $D$.
It is a highly efficient algorithm for fully connected conditional random field models, in which the pairwise edge potentials between all pairs of pixels in the image are defined by a linear combination of Gaussian kernels in a arbitrary feature space.
The iterative optimisation of the random field requires the propagation of all the potentials across the image.
The high efficiency of this method relies on approximating the random field with a mean field whose update can be computed using Gaussian filtering in the feature space, reducing the computational complexity from quadratic to linear in the number of pixels.
The user only needs to set the number of iterations and the kernel width of the Gaussian kernels.
We set the former equal to $5$ and the latter equal to $0.1$.

The change map can be obtained by thresholding the filtered version of $\boldsymbol{D}$.
Assuming that the image can be segmented into foreground (changes) and background (no changes), the best threshold is sought out to split the image histogram in two parts.
The optimal threshold can be determined heuristically or by automatic thresholding methods such as~\cite{kapur1985new,shanbhag1994utilization,yen1995new}, while we have used the classical Otsu's method~\cite{otsu1979threshold}.
It is a well-known method widely used in the image processing literature to find the optimal threshold for two-class segmentation of greyscale images exhibiting bimodal histograms.
In a fashion which recalls Fisher's discriminant analysis, Otsu's method picks up the threshold value that minimises the intra-class variances and maximises the inter-class variance.

In the following we briefly describe the regression methods considered in this work to evaluate $f^{(1)}$ and $f^{(2)}$.

%
\subsubsection{Gaussian Process Regression}
A Gaussian process (GP) is a collection of random variables, $\{\boldsymbol{x}_i\}$, taking values in $\mathbf{R}^P$, any finite subset of which has a joint multivariate Gaussian distribution.
It is completely specified by its mean function $\boldsymbol{m}\left(\boldsymbol{x}\right)$ and covariance (kernel) function $k_{\boldsymbol{x}_i,\boldsymbol{x}_j}=k\left(\boldsymbol{x}_i,\boldsymbol{x}_j\right)$.
For regression purposes, zero mean GPs are most often used~\cite{rasmussen2004gaussian}.
Given a training set of $M$ input vectors arranged in rows, $\boldsymbol{X}\! \in\! \mathbf{R}^{M\! \times\! P}$, a corresponding set of target vectors, $\boldsymbol{Y}\! \in\! \mathbf{R}^{M\! \times\! Q}$, and a set of $N$ new observed vectors $\boldsymbol{X}_*\! \in\! \mathbf{R}^{N\! \times\! P}$, the joint distribution of the training vectors $\boldsymbol{Y}$ and the sought vectors $\boldsymbol{Y}_*\! \in\! \mathbf{R}^{N\! \times\! Q}$, conditioned on the input data $\boldsymbol{X}$ and $\boldsymbol{X}_*$, is
\begin{equation}
\left[ \begin{tabular}{c}
    $\boldsymbol{Y}$ \\
    $\boldsymbol{Y_*}$
\end{tabular}\right] \arrowvert\boldsymbol{X},\boldsymbol{X}_*
\sim \mathcal{N} \left(\boldsymbol{0},\left[\begin{tabular}{c c}
    $\boldsymbol{K}_{\boldsymbol{X},\boldsymbol{X}}$ &  $\boldsymbol{K}_{\boldsymbol{X},\boldsymbol{X_*}}$\\
    $\boldsymbol{K}_{\boldsymbol{X_*},\boldsymbol{X}}$ &  $\boldsymbol{K}_{\boldsymbol{X_*},\boldsymbol{X_*}}$
\end{tabular}\right]\right).
 \end{equation}
where $\boldsymbol{K}_{\boldsymbol{X},\boldsymbol{X_*}}$ is the matrix whose entry $(i,j)$ is the covariance between the $i$th row of $\boldsymbol{X}$ and the $j$th row of $\boldsymbol{X}_*$. Matrices $\boldsymbol{K}_{\boldsymbol{X},\boldsymbol{X}}$, $\boldsymbol{K}_{\boldsymbol{X_*},\boldsymbol{X}}={\boldsymbol{K}_{\boldsymbol{X},\boldsymbol{X_*}}}^T$, and $\boldsymbol{K}_{\boldsymbol{X_*},\boldsymbol{X_*}}$ have analogue definitions.
Thus, the following posterior distribution is derived (see~\cite{rasmussen2004gaussian} for details):
\begin{equation}
\begin{split}
\boldsymbol{Y_*}\! \mid\! \boldsymbol{X_*},\boldsymbol{X},\boldsymbol{Y}\! &\sim\! \mathcal{N}\! \big(\!\boldsymbol{K}_{\boldsymbol{X_*},\boldsymbol{X}}\cdot\boldsymbol{K}_{\boldsymbol{X},\boldsymbol{X}}^{-1}\cdot\boldsymbol{Y}, \\ & \boldsymbol{K}_{\boldsymbol{X_*},\boldsymbol{X_*}}-\boldsymbol{K}_{\boldsymbol{X_*},\boldsymbol{X}}\cdot\boldsymbol{K}_{\boldsymbol{X},\boldsymbol{X}}^{-1}\cdot\boldsymbol{K}_{\boldsymbol{X},\boldsymbol{X_*}}\big)
\end{split}
 \end{equation}
Hence, the corresponding conditional mean is the maximum {\it a-posteriori} prediction $\boldsymbol{\hat{Y}}=\boldsymbol{K}_{\boldsymbol{X_*},\boldsymbol{X}}\cdot\boldsymbol{K}_{\boldsymbol{X},\boldsymbol{X}}^{-1}\cdot\boldsymbol{Y}$.
The two main factors affecting the quality of the regression are the choice of kernel function and its hyperparameters.
In this work, we opted for the commonly used radial basis function (RBF)
 \begin{equation}
k_{\boldsymbol{x}_i,\boldsymbol{x}_j}=\sigma_f^2\exp\left(-\frac{1}{2}\left(\boldsymbol{x}_i-\boldsymbol{x}_j\right)^TL\left(\boldsymbol{x}_i-\boldsymbol{x}_j\right)\right)\,,
 \end{equation}
where $\boldsymbol{\theta} = \left\{L,\sigma_f^2\right\}$ is the set of hyperparameters, with signal variance $\sigma_f^2$ and $L = \ell^{-2}I$, if the length-scale parameter $\ell$ is a scalar (isotropic kernel), or $L = \text{diag}\left(\boldsymbol{\ell}^{-2}\right)$, if $\boldsymbol{\ell}$ is a vector (anisotropic kernel)~\cite{rasmussen2004gaussian}.
Concerning the optimisation of $\boldsymbol{\theta}$, a gradient ascent is performed to maximise the marginal likelihood $\mathcal{P}\left(\boldsymbol{Y}\mid\boldsymbol{X},\boldsymbol{\theta}\right)$.
%
%
A weak point of this optimisation is that it might lead to a local maximum instead of the global one, so it is recommended to iterate the procedure several times starting from different random points in the hyperparameter space $\Omega_{\boldsymbol{\theta}}$.


\subsubsection{Multiple-output Support Vector Regression}

Support vector machines (SVMs) are well-known machine learning algorithms used for classification and regression.
By solving the so-called dual problem~\cite{cortes1995support}, it is possible to find the best separating or fitting curve with respect to a loss function that accounts for misclassification or reconstruction error and with respect to a regularisation parameter which defines the width of a soft margin around such a curve.
In addition, the support vectors, i.e.\ the training points that define the 
margin, are highlighted from the rest of the training set.

Instead of coping with a multiple-output problem (i.e.\ the regression of a multivariate variable) all at once, the solution usually adopted is to train a dedicated SVM for each regressand variable.
Therefore, the standard implementations of SVR are designed to predict a single output feature, ignoring the potentially nonlinear relations across the target features~\cite{tuia2011multioutput}.
Tuia et al.~\cite{tuia2011multioutput} proposed a multiple-input multiple-output SVR method to overcome this limitation.
During the training phase, it aims to minimise the cost function
 \begin{equation}
L_\text{SVR}\left(\boldsymbol{W},\boldsymbol{b}\right) = \frac{1}{2}\sum_{q=1}^Q\parallel\boldsymbol{w}_q\parallel^2+\lambda\sum_{m=1}^M L\left(\mu_m\right)
 \end{equation}
where
\begin{align}
\label{eq:errorterms}
 L \left(\mu_m\right) & =
\begin{cases}
    0 & \mu_m < \epsilon \\
    \mu_m^2-2\mu_m\epsilon+\epsilon^2 & \mu_m \geq \epsilon
\end{cases} \ , \\
 \mu_m & = \ \parallel\boldsymbol{e}_m\parallel \ =\sqrt{\boldsymbol{e}_m^T\boldsymbol{e}_m} \ , \\
 \boldsymbol{e}_m^T & = \boldsymbol{y}_m^T\! -\phi\! \left(\boldsymbol{x}_m\right)^T\boldsymbol{W}\! -\boldsymbol{b}^T.
\end{align}
Here, $\boldsymbol{W}=\left[\boldsymbol{w}_1,\dots,\boldsymbol{w}_Q\right]$ with $\boldsymbol{w}_q\in \mathbf{R}^{P'}$ are the coefficients and $\boldsymbol{b}=\left[b_1,\dots,b_Q\right]^T$ are the biases in the linear combination of the data points $\boldsymbol{x}_m$ transferred into a finite-dimensional space by the kernel function $\phi:\mathbf{R}^P\longrightarrow\mathbf{R}^{P'}$.
The extension to a (possibly infinitely-dimensional) separable Hilbert space is straightforward.
We recall that $M$ is the number of training data points, and $P$ ($Q$) is the number of features at time $t_1$ ($t_2$).
The penalty factor $\lambda$ sets the trade-off between the regularisation term and the sum of the error terms $L\left(\mu_m\right)$.
If $\lambda$ is too large, nonseparable points would highly penalise the cost function and too many data points will turn into support vectors, favoring overfitting.
Vice versa, a small $\lambda$ may lead to underfitting.
The parameter $\epsilon$ is half the width of the insensitivity zone.
This zone delimits a "tube" around the approximated function and the training data points within this insensitivity zone do not contribute to the cost function (see Eq.\ \ref{eq:errorterms}).
For too small values of $\epsilon$, too many data points will be considered as support vectors (overfitting), the generalisation performance will be affected and the variance of the fitted curve will be too large.
On the contrary, a too large $\epsilon$ will cause underfitting and the overall accuracy will be low.
Another critical hyperparameter is the width $\sigma$ of the RBF kernel $\phi$.
To select the right combination of hyperparameters $\boldsymbol{\theta} = \left\{\lambda,\epsilon,\sigma\right\}$, a grid search for the smallest cross-validation error or the minimization of an error bound can be applied.
Once the optimisation of the parameters is performed, the regression consists of
\begin{equation}\label{eq:svm_predictor}
    \boldsymbol{\hat{y}_*}= \boldsymbol{W}_{opt}^T\, \phi\left(\boldsymbol{x}_*\right)\! +\boldsymbol{b}_{opt}.
\end{equation}


\subsubsection{Random Forest Regression}

The random forest (RF) approach was proposed by Breiman in~\cite{breiman2001random} to perform both classification and regression by exploiting the simplicity of decision trees and the robustness of ensemble methods.
From now on, only regression will be considered, but for classification purposes the approach is similar.

A RF consists of $T$ trees, at whose nodes $r$ randomly selected features are compared to thresholds (e.g., $\text{feat}_1 > \text{thr}_1 \ \& \dots \& \ \text{feat}_r > \text{thr}_r)$.
These thresholds are determined during the training of the trees, for which various algorithms (e.g., classification and regression tree~\cite{breiman2017classification}) have been developed.
In each tree, the training data points are divided over the branches according to these conditions, and the trees expand until only one data point or a predefined maximum number of data points (e.g., 5) is contained in each of the final nodes (leaves).
Thus, the average of the corresponding training vectors $\boldsymbol{y}_m$ is assigned to each leaf.
During the test phase an input vector $\boldsymbol{x_*}$ propagates through each tree and reaches one of the leaves, giving as the output $\boldsymbol{y}_t$ the aforementioned assigned value.
Finally, the average of the $T$ outputs, $\boldsymbol{y}_t$, is computed, thereby obtaining the final regressed vector $\boldsymbol{\hat{y}_*}$:

\begin{equation}
   \boldsymbol{\hat{y}_*} = \frac{1}{T}\sum_{t=1}^T \boldsymbol{y}_t
\end{equation}

To generalise better, every tree is trained on a bootstrap sample drawn from the training set, and a randomly drawn subset of features (of fixed cardinality) is used on each node of each tree.
The validation is carried out through out-of-bag estimation~\cite{breiman2001random}.
Moreover, the behaviour of a RF can be controlled by tuning three parameters: the size of the forest (i.e.\ the number of trees $T$), the number of features $r$ considered in every node, and the depth of the trees.

Concerning the number of features considered at every node, in~\cite{breiman2001random} it is suggested by empirical results to set $m=\lfloor\frac{\log P}{\log 2}\rfloor$, where $P$ is the dimension of the vectors $\boldsymbol{x}$.
Another rule of thumb is to use: $m=\lfloor\nicefrac{P}{3}\rfloor$~\cite{breiman2001random}.
However, there are no practical rules to choose the size of the forest.
One may think that for a larger number of trees the outcomes become better, but~\cite{breiman2001random} proved that at some point the overall accuracy saturates due to correlation between the trees.
Therefore, a compromise between gained accuracy and computational load must be found.


\subsubsection{Homogeneous Pixel Transformation}

The homogeneous pixel transformation (HPT) method proposed by Liu et al.~\cite{liu2018change} is a kernel regression algorithm~\cite{wand1995kernel} based on the $K$-nearest neighbours of each data point.
This technique recalls the distance weighted averaging or locally weighted regression previously presented in~\cite{atkeson1997locally}, where many related aspects are also studied: possible kernels, distance measures, choices of the bandwidth, denoising techniques, and outlier detection.
For every data point in the first image, $\boldsymbol{x}_i$, the $K$ nearest neighbours among the training vectors $\boldsymbol{x}_m \in \mathcal{T}$ are sought for.
The regression consists of the weighted sum
 \begin{equation}
 \boldsymbol{\hat{y}}_i = \sum_{k=1}^{K}w_{i,k}\cdot \boldsymbol{y}_{i,k}\ ,
 \end{equation}
 where
  \begin{equation}
 w_{i,k} = w\left(\boldsymbol{x}_i,\boldsymbol{x}_k\right) = e^{-\gamma d_{i,k}}\ .
 \end{equation}
Here $d_{i,k}$ is the Euclidean distance between $\boldsymbol{x}_i$ and its $k^{th}$ nearest neighbour $\boldsymbol{x}_k$, $\boldsymbol{y}_{i,k}$ is the corresponding vector of $\boldsymbol{x}_k$ in $\mathcal{T}$, whereas the kernel width $\gamma$ regulates how strongly the farthest neighbours are penalised.
If $\gamma$ is too small, the addends tend to be equally weighted and the sum is close to an average, if $\gamma$ is too large, few main addends contribute to the sum whilst the rest are heavily penalised.
Before computing the weights, a relative normalisation of the distances is applied:
 \begin{equation}
 d_{i,k} = \frac{\left\Arrowvert\boldsymbol{x}_i-\boldsymbol{x}_k\right\Arrowvert}{\max_k\left\Arrowvert\boldsymbol{x}_i-\boldsymbol{x}_k\right\Arrowvert}\ .
 \end{equation}
The normalisation in~\cite{liu2018change} is defined as relative, because it considers the maximum among the distances between the data point $\boldsymbol{x}_i$ and its neighbours.
However, while testing our implementation, we found that it is better to perform an absolute normalisation, thus seeking the maximum among all the computed distances.

\subsection{Methodological comparison among the considered regression methods}
A common trait shared by all aforementioned regression methods is their fully nonparametric formulation. From a methodological perspective, this property makes them applicable to input datasets $(\boldsymbol{X},\boldsymbol{Y})$ with arbitrary probability distribution.

Indeed, the ensemble formulation of RF, in which bagging is paired with random feature selection, results in a remarkable robustness to overfitting~\cite{breiman2001random,Merentitis}. Similarly, the SVM predictor (\ref{eq:svm_predictor}) is proven to be a kernel expansion in which kernels are centered on a (usually small) subset of automatically selected training samples (the well-known support vectors)~\cite{Vapnik1998}. Hence, the regression solution is generally sparse, which, in turn, yields robustness to overfitting. The GPR predictor can also be expressed as an expansion of autocovariance functions~\cite{rasmussen2004gaussian}. On one hand, this bears formal similarities to the SVM expansion, especially because if the autocovariance of a GP is a continuous function, it also is a legit kernel function (i.e., its value is equivalent to computing an inner product in some transformed Hilbert space)~\cite{ash1990information}. On the other hand, the GPR expansion is generally dense (i.e., autocovariance terms are centered on all training samples), a property that may cause sensitivity to overfitting. Similar to the case of $K$-NN classification, HPT may also be sensitive to overfitting when $K$ is too small.

The difference between the kernel expansions of SVM and GPR also impacts on the computational complexity of the prediction phase, which is linear in the number of support vectors for SVM and in the total number of training samples for GPR. RF usually exhibits a low computational burden that grows linearly with the number of trees in the forest~\cite{breiman2001random,Merentitis}. In the case of HPT, limited computational burden may be ensured by making use of $K$-d tree formulations~\cite{kdtree}.

As the SVM and GPR predictors are kernel expansions, they are smooth functions as long as the related kernels are smooth. The HPT predictor is also generally regular. On the contrary, the predictor determined by a regression tree is piecewise constant in the feature space~\cite{breiman2017classification}. As the RF predictor is the average of the outputs of the corresponding trees, it also is piecewise constant. In general, this behavior may be undesired when a smooth predictor is sought for. Within the proposed change detection method, this aspect does not represent a limitation, though.

In terms of model selection, the Bayesian rationale of GPR naturally endows it with maximum likelihood-type methods for the optimization of the autocovariance hyperparameters~\cite{rasmussen2004gaussian}. In the case of SVM, this property does not hold intrinsically, but several case-specific algorithms have been proposed for hyperparameter optimization based, for instance, on minimizing generalization error bounds~\cite{Chapelle,moser_serpico2009}. The performance of RF usually exhibits limited sensitivity to its own hyperparameters, thus making model selection quite straightforward~\cite{Merentitis}. Furthermore, an error measure intrinsic to the RF process, the so-called out-of-bag error, can also be used to guide the tuning of the hyperparameters~\cite{breiman2001random,Merentitis}. In the case of HPT, guidelines on hyperparameter tuning are provided in~\cite{liu2018change} although automatic hyperparameter-optimization methods have not been developed so far. A general strategy may be based on cross-validation, although at the cost of increased computation time.

\section{Experimental results}\label{sec:results}

The self-supervised training set selection is carried out on two different datasets.
There are many different kinds of terrain involved, and excluding any of them might lead to poor results.
The investigated sizes $M$ of the set $\mathcal{T}$ are $M=10^2,10^3,10^4,10^5$.
The largest $M$ corresponds to $8.07\%$ of $N$ for the first case study and $1.43\%$ of $N$ for the second case study.
The unsupervised training data selection based on affinity matrix distances is applied with patch sizes of $k = 5, 10, 20$.
The experiments were performed on a machine running Ubuntu 14 with a 8-core CPU @ $2.7$ GHz and $64$ GB of RAM, using all cores to exploit the potential for parallel processing in the various methods.

The performance of the proposed CD framework configured with the different regression methods is evaluated in terms of accuracy and computational speed.
We measure the quality of the result, both after the affinity matrices comparison and the regression, in terms of \emph{area under the curve} (AUC), with values ranging between $0.5$ (poor) and $1$ (optimal), or equivalently between $50\%$ and $100\%$.
This indicates the area under the receiver (i.e. detector) operating characteristic curve, which measures the false positive rate against the true positive rate.
The change detection accuracy is also evaluated in terms of overall accuracy (OA), which is the ratio of correctly classified data points divided by the total number of data points $N$, and Cohen's Kappa Coefficient (KC) \cite{cohen1960coefficient}, which is expressed as 
\begin{equation}
    \text{KC} = \frac{p_o - p_e}{1 - p_e}.
\end{equation}
Here $p_o$ is the observed agreement between predictions and labels, i.e.\ the OA, while
$p_e$ is the probability of random agreement, which is estimated from the observed true positives (TP), true negatives (TN), false positives (FP), and false negatives (FN) as:
\begin{equation}
\begin{split}
    p_e = & \left(\frac{\text{TP} + \text{FP}}{N} \cdot \frac{\text{FN} + \text{TN}}{N}\right) \\  + & \left(\frac{\text{TP} + \text{FN}}{N} \cdot \frac{\text{FP} + \text{TN}}{N}\right)\,.
\end{split}
\end{equation}
Computation speed is measured as the elapsed time during computation of the regression analysis in both directions, starting from the training phase and ending after the test phase.
It must be pointed out that two of the methods are implemented in Python libraries (GPR and RFR), whereas the code provided by \cite{tuia2011multioutput} for the MIMO SVR method is written in MATLAB, and so is our implementation of the HPT method.
Therefore, an exact comparison of execution time of each algorithm is not possible, even though the two programming languages are both interpreted.
Nevertheless, the run times are indicators that can help us rank the four algorithms in terms of speed.

\subsection{Forest fire in Texas}

\begin{figure}[ht!]

\begin{subfigure}[t]{0.3\columnwidth}
\includegraphics[width=\linewidth,keepaspectratio]{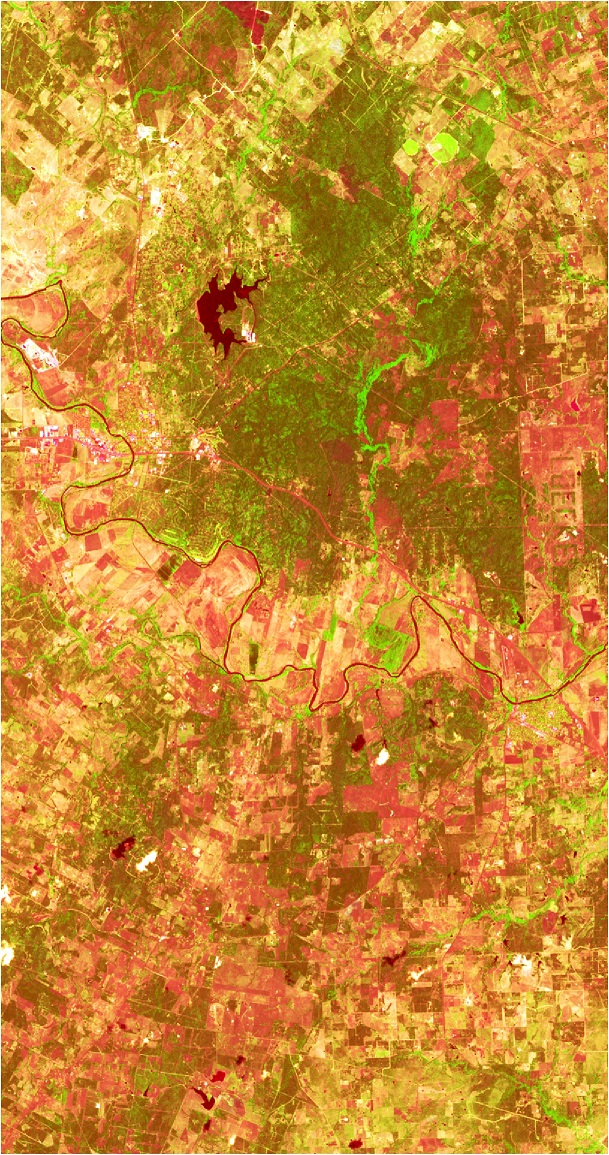}
\caption{Landsat 5 ($t_1$)}
\label{fig:L5}
\end{subfigure}
\hfill
\begin{subfigure}[t]{0.3\columnwidth}
\includegraphics[width=\linewidth,keepaspectratio]{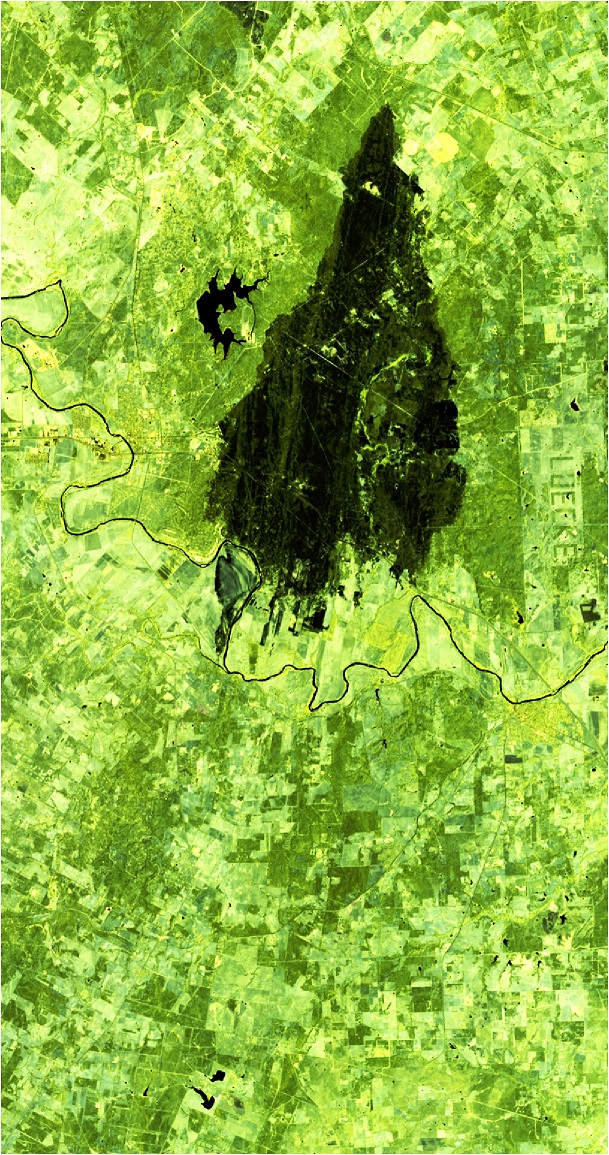}
\caption{EO-1 ALI ($t_2$)}
\label{fig:ALI}
\end{subfigure}
\hfill
\begin{subfigure}[t]{0.3\columnwidth}
\includegraphics[width=\linewidth,keepaspectratio]{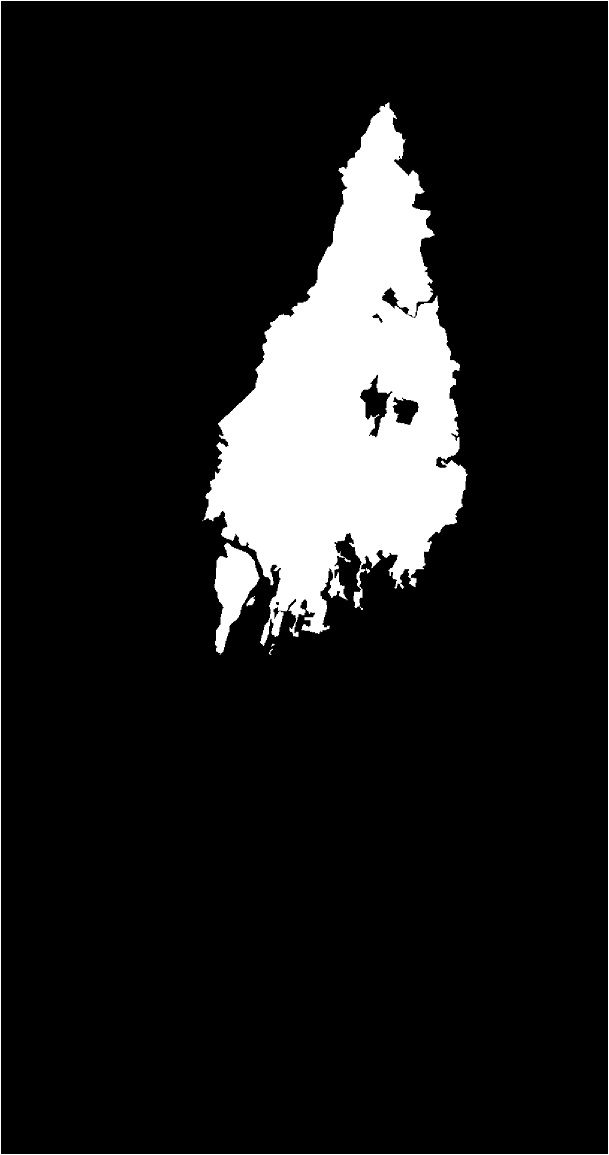}
\caption{Ground Truth}
\label{fig:gt1}
\end{subfigure}
\caption{Forest fire in Texas: Landsat 5 ($t1$), (b) EO-1 ALI ($t2$), (c) ground truth. RGB false color composites are shown for both images.}
\label{fig:dataset1}
\end{figure}

The first dataset is composed of a multispectral $1534 \times 808$ image acquired by Landsat 5 TM (Fig.\ \ref{fig:L5}) before a forest fire in Bastrop County, Texas, during September-October, 2011\footnote{Distributed by LP DAAC, http://lpdaac.usgs.gov \label{foot1}}.
An Earth Observing-1 Advanced Land Imager (EO-1 ALI) multispectral acquisition after the event completes the dataset (Fig. \ref{fig:ALI})$^1$.
Both images are optical with $7$ and $10$ channels, respectively, some of which cover the same spectral bands, so the signatures of the classes involved are partly similar.
Among the possible heterogeneous CD scenarios, this is one of the easiest.
The ground truth of the event (see Fig.\ \ref{fig:gt1}) is provided by Volpi et al.\ \cite{volpi2015spectral}.

\subsubsection{Training set selection}
\begin{figure}[ht!]
\begin{center}
\begin{subfigure}[t]{0.4\columnwidth}
\includegraphics[width=\linewidth,keepaspectratio]{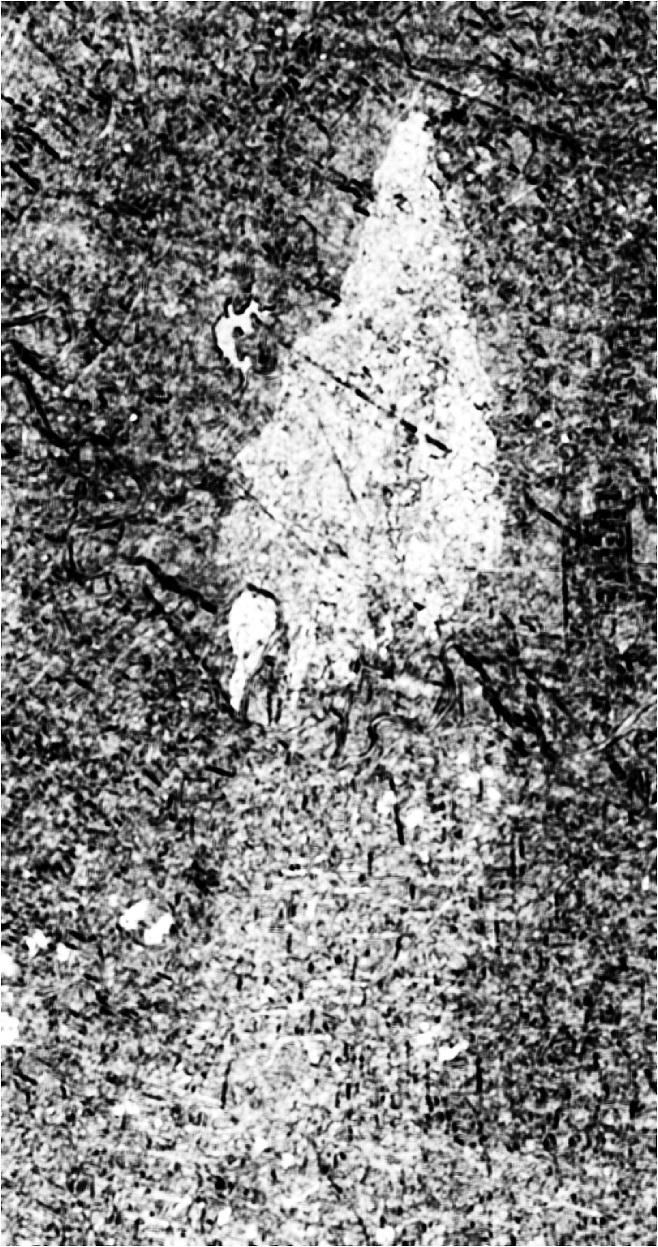} 
\caption{\centering Heat-map,\newline $k=5$}
\label{fig:hm5_tx}
\end{subfigure}
\hspace{0.05\columnwidth}%
\begin{subfigure}[t]{0.4\columnwidth}
\includegraphics[width=\linewidth,keepaspectratio]{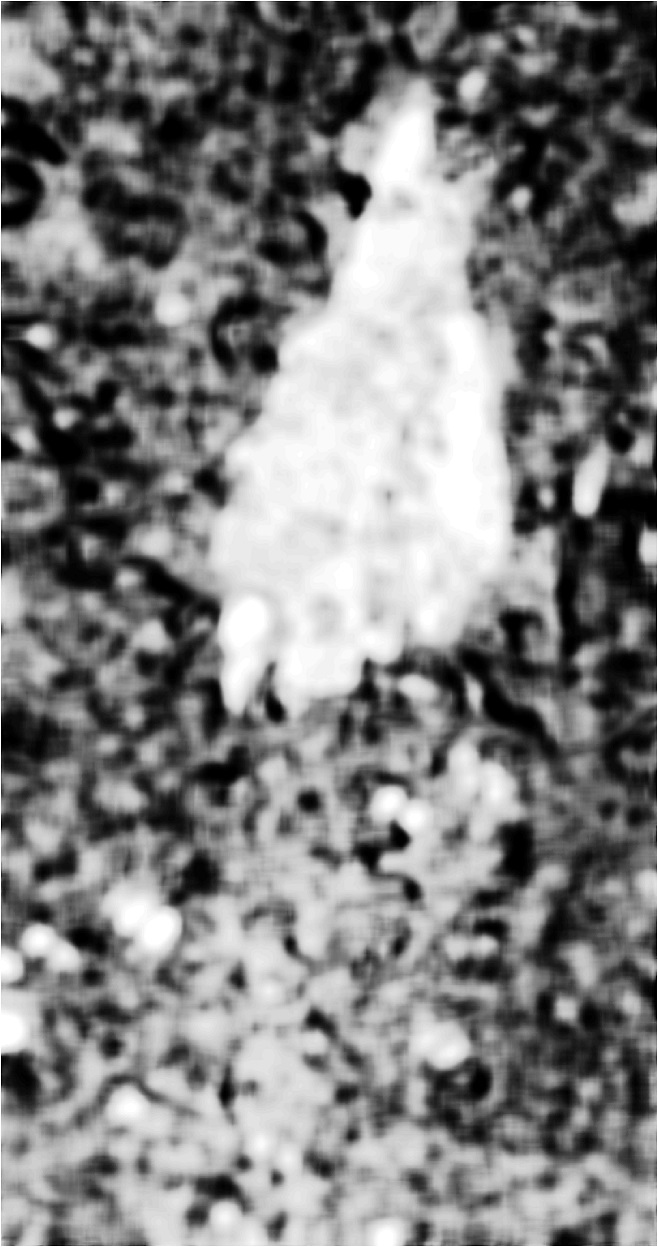}
\caption{\centering Heat-map,\newline $k=20$}
\label{fig:hm20_tx}
\end{subfigure}

\end{center}

\caption{Affinity matrices comparison method for training set selection: heat-maps for $k=5$ (a) and $k=20$ (b).}
\label{fig:affm_tx}
\end{figure}

\begin{figure}[ht!]
\begin{center}
\begin{subfigure}[t]{0.49\columnwidth}
\includegraphics[width=\linewidth,keepaspectratio]{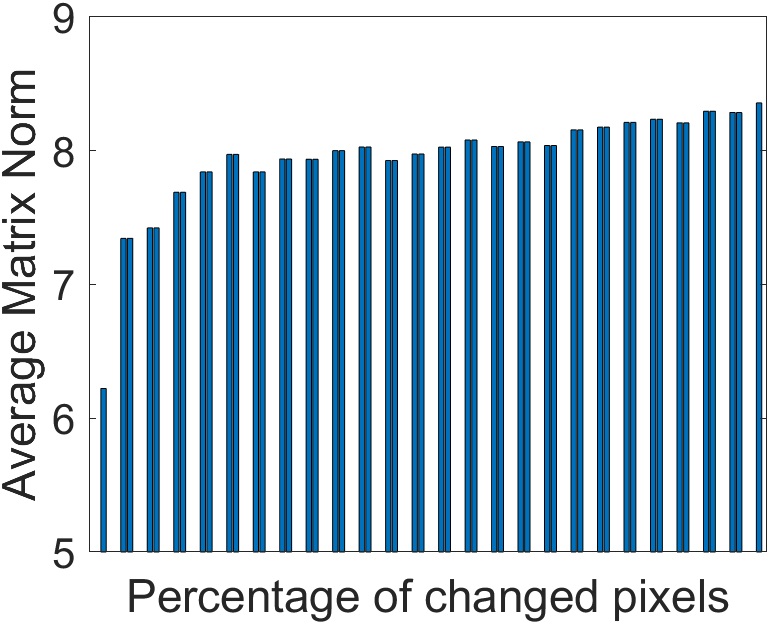}
\caption{$k=5$}
\label{fig:mean_f_tx_5}
\end{subfigure}~%
\begin{subfigure}[t]{0.49\columnwidth}
\includegraphics[width=\linewidth,keepaspectratio]{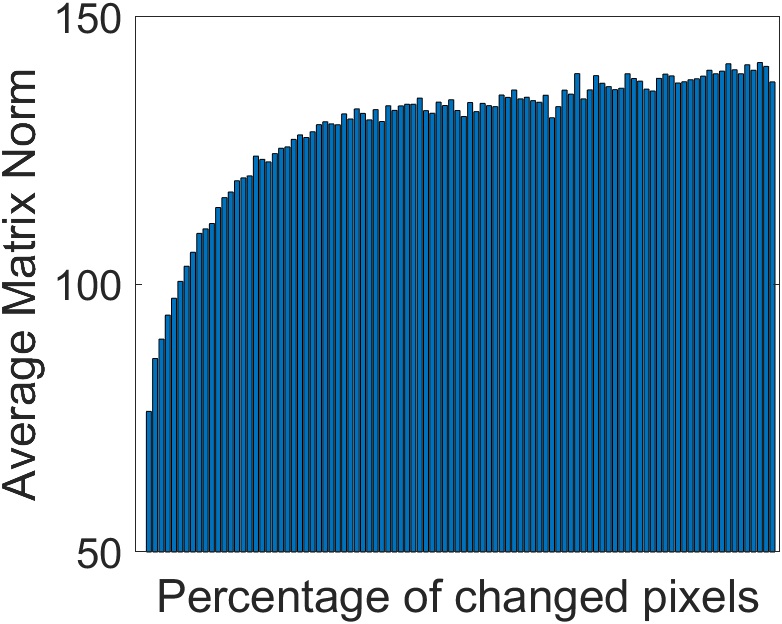}
\caption{$k=20$}
\label{fig:mean_f_tx_20}
\end{subfigure}
\end{center}
\caption{Average matrix norm $f$ versus the percentage of changed pixels, for $k=5$ (a) and $k=20$ (b).}
\label{fig:mean_f_tx}
\end{figure}

For patch sizes $k=5$ and $k=20$, the heat-maps for the average matrix norm are shown in Fig.\ \ref{fig:hm5_tx} and \ref{fig:hm20_tx}, respectively;
It is worth underlining that even by thresholding the obtained heat-map we could obtain an already reasonable change map.
More specifically, we achieve an AUC equal to $84.9\%$, $91.3\%$, and $93.1\%$ for $k=5,10,20$ respectively.
We also report that the whole procedure took about $1$ minute, $5$ minutes and $20$ minutes for $k=5,10,20$ respectively.
To evaluate the hypothesis that a patch $p$ associated to a small matrix norm $f$ is most likely to cover an unchanged area, we use the ground truth as follows. For each patch $p$, the pixels belonging to changed areas are counted, so each patch $p$ goes from having $0\%$ of its pixels affected by changes to $100\%$ of them.
Fig.\ \ref{fig:mean_f_tx} illustrates the average matrix norm $f$ with respect to the percentage of changed pixels, for $k=5$ and $k=20$.
The monotonic trend for which an increasing number of changed pixels leads on average to a larger matrix norm can be already noticed for $k=5$, and it is evident for $k=20$.

\begin{table}[ht!]
\centering
\setlength{\tabcolsep}{4pt}
\small
\caption{AUC, FN, $d_H\left(\mathcal{H}_{\boldsymbol{X}},\mathcal{H}_{\boldsymbol{X} \cap \mathcal{T}}\right)$ and $d_H\left(\mathcal{H}_{\boldsymbol{Y}},\mathcal{H}_{\boldsymbol{Y} \cap \mathcal{T}}\right)$ for various combinations of $k$ and $M$ on the Texas dataset.}
\label{tab:Aff_Tx}

\begin{tabular}{ c c | c c c}
\toprule
\multicolumn{2}{c|}{Patch size $k$} & $k = 5$ & $k = 10$ & $k = 20$ \\
\midrule
 \multicolumn{2}{c|}{AUC} & 84.9 & 91.3 & 93.1\\
 \hline
\multirow{4}{*}{FN} & $M = 10^2$ & 0 & 0 & 0 \\
& $M = 10^3$ & 0 & 0 & 0 \\
& $M = 10^4$ & 0.150 & 0.010 & 0 \\
& $M = 10^5$ & 0.312 & 0.070 & 0 \\
\hline
\multirow{4}{*}{$d_H\left(\mathcal{H}_{\boldsymbol{X}},\mathcal{H}_{\boldsymbol{X} \cap \mathcal{T}}\right)$} & $M = 10^2$ & 0.416 & 0.448 & 0.457 \\
& $M = 10^3$ & 0.236 & 0.235 & 0.226 \\
& $M = 10^4$ & 0.149 & 0.140 & 0.158 \\
& $M = 10^5$ & 0.133 & 0.107 & 0.010 \\
\hline
\multirow{4}{*}{$d_H\left(\mathcal{H}_{\boldsymbol{Y}},\mathcal{H}_{\boldsymbol{Y} \cap \mathcal{T}}\right)$} & $M = 10^2$ & 0.628 & 0.641 & 0.656 \\
& $M = 10^3$ & 0.299 & 0.294 & 0.315 \\
& $M = 10^4$ & 0.155 & 0.162 & 0.185 \\
& $M = 10^5$ & 0.132 & 0.119 & 0.121 \\
\bottomrule
\end{tabular}

\end{table}

\begin{figure*}[b!]

\begin{center}
\begin{subfigure}[t]{0.49\columnwidth}
\includegraphics[width=\linewidth,keepaspectratio]{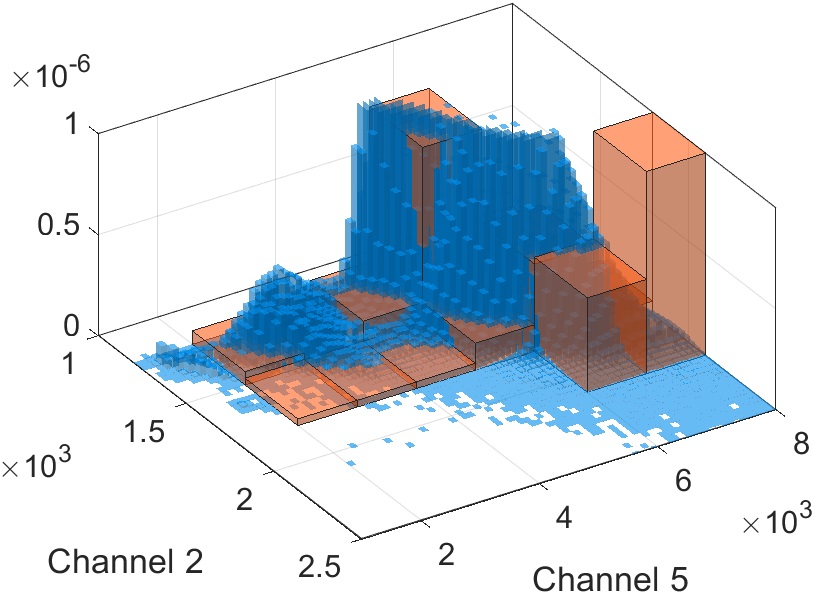}
\caption{\centering $\mathcal{H}_{\boldsymbol{Y}}$ (blue), $\mathcal{H}_{\boldsymbol{Y} \cap \mathcal{T}}$ (red) \newline $M=10^2$}
\label{fig:hist_tx_1}
\end{subfigure}
\begin{subfigure}[t]{0.49\columnwidth}
\includegraphics[width=\linewidth,keepaspectratio]{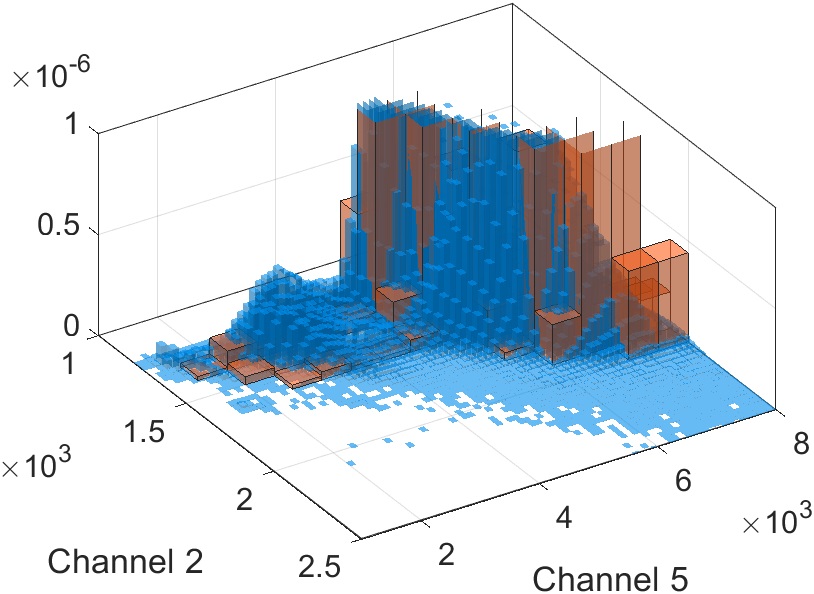}
\caption{\centering $\mathcal{H}_{\boldsymbol{Y}}$ (blue), $\mathcal{H}_{\boldsymbol{Y} \cap \mathcal{T}}$ (red) \newline $M=10^3$}
\label{fig:hist_tx_2}
\end{subfigure}
\begin{subfigure}[t]{0.49\columnwidth}
\includegraphics[width=\linewidth,keepaspectratio]{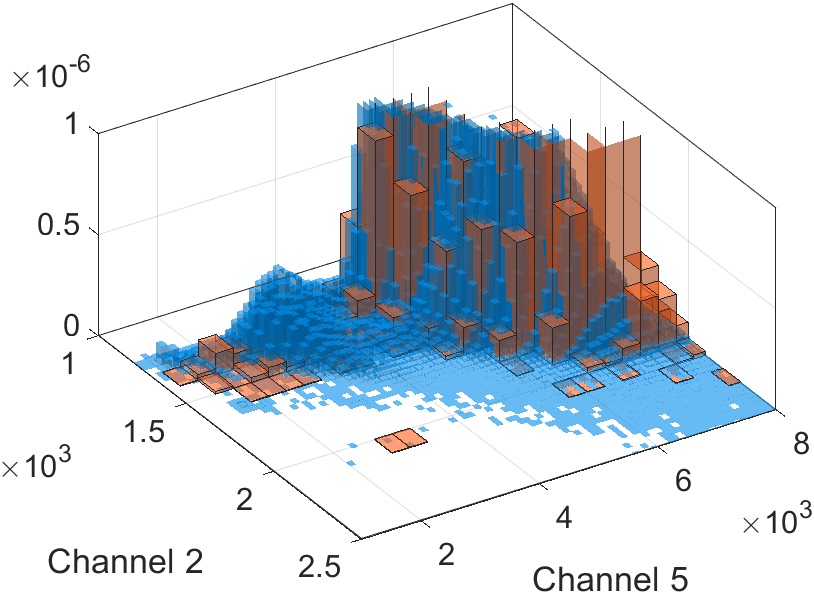}
\caption{\centering $\mathcal{H}_{\boldsymbol{Y}}$ (blue), $\mathcal{H}_{\boldsymbol{Y} \cap \mathcal{T}}$ (red) \newline $M=10^4$}
\label{fig:hist_tx_3}
\end{subfigure}
\begin{subfigure}[t]{0.49\columnwidth}
\includegraphics[width=\linewidth,keepaspectratio]{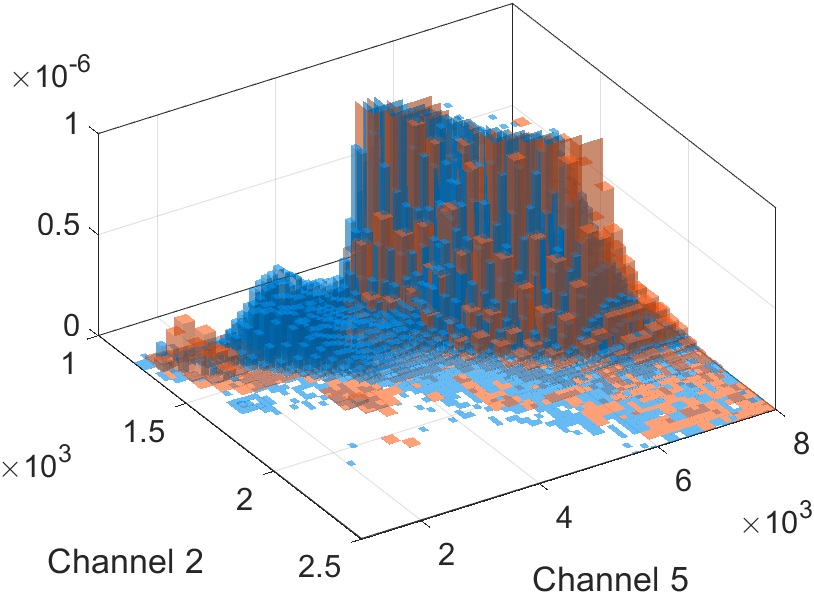}
\caption{\centering $\mathcal{H}_{\boldsymbol{Y}}$ (blue), $\mathcal{H}_{\boldsymbol{Y} \cap \mathcal{T}}$ (red) \newline $M=10^5$}
\label{fig:hist_tx_4}
\end{subfigure}
\end{center}

\caption{Example of comparison between $\mathcal{H}_{\boldsymbol{Y}}$ and $\mathcal{H}_{\boldsymbol{Y} \cap \mathcal{T}}$ with two channels of $\boldsymbol{Y}$ for $k=20$: $\mathcal{H}_{\boldsymbol{Y}}$ is depicted in blue, whereas $\mathcal{H}_{\boldsymbol{Y} \cap \mathcal{T}}$ is depicted in red for $M = 10^2,10^3,10^4,10^5$ in (a), (b), (c) and (d) respectively. Best viewed in colour.}

\label{fig:hist_tx}
\end{figure*}

Table \ref{tab:Aff_Tx} summarises the AUC, the Hellinger distances and the percentage of changed pixels included in $\mathcal{T}$, denoted as FN, for the considered values of $k$ and $M$.
We point out that the method largely succeeds in avoiding changed pixels across all the cases shown in Table \ref{tab:Aff_Tx}.
Having a larger number of data points in $\mathcal{T}$ slightly increases the chances of including changed pixels, but the false negatives are still negligable with respect to the total size of $\mathcal{T}$, and go to zero as the patch size $k$ becomes larger.
Regarding the Hellinger distances, they are quite high for $M = 10^2$, but otherwise limited to reasonable values.
Moreover, two general trends can be noticed.
First, $d_H\left(\mathcal{H}_{\boldsymbol{X}},\mathcal{H}_{\boldsymbol{X} \cap \mathcal{T}}\right)$ is consistently smaller than $d_H\left(\mathcal{H}_{\boldsymbol{Y}},\mathcal{H}_{\boldsymbol{Y} \cap \mathcal{T}}\right)$.
Keeping in mind that for this dataset the changes are represented by a specific class of $\boldsymbol{Y}$ (forest fire scar), it is reasonable to infer that the method is excluding these pixels from $\mathcal{T}$, giving rise to a bigger discrepancy between $\mathcal{H}_{\boldsymbol{Y}}$ and $\mathcal{H}_{\boldsymbol{Y} \cap \mathcal{T}}$, and consequently to a larger Hellinger distance.
Second, the distances tend to 0 as more training points are considered.
This does not surprise: Although few data points might cover most of the image domain and all the classes involved in modality $\boldsymbol{l}$, they are not enough to yield a good estimate the latent data distribution.

\begin{figure}[ht!]
\begin{center}
\begin{subfigure}[t]{0.4\columnwidth}
\includegraphics[width=\linewidth,keepaspectratio]{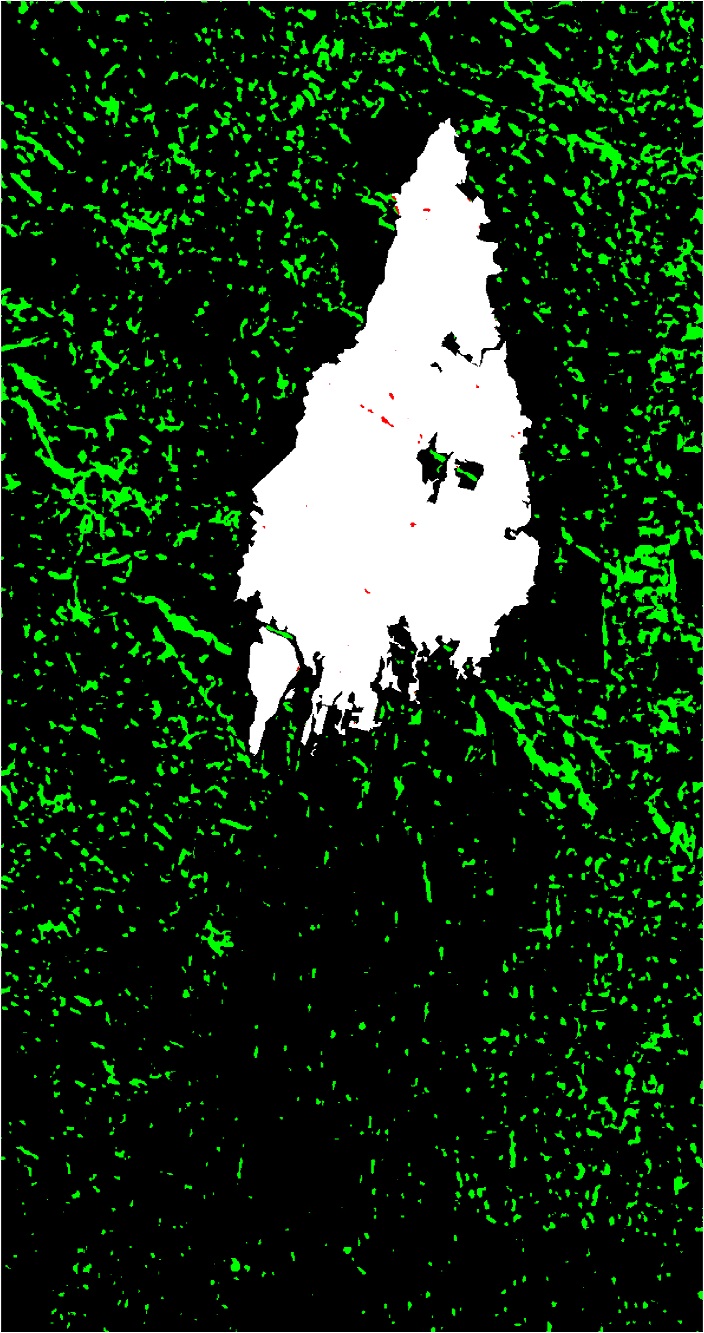}
\caption{\centering Training set, \newline $M = 10^5$, $k=5$}
\label{fig:ts5_tx}
\end{subfigure}
\hspace{0.05\columnwidth}%
\begin{subfigure}[t]{0.4\columnwidth}
\includegraphics[width=\linewidth,keepaspectratio]{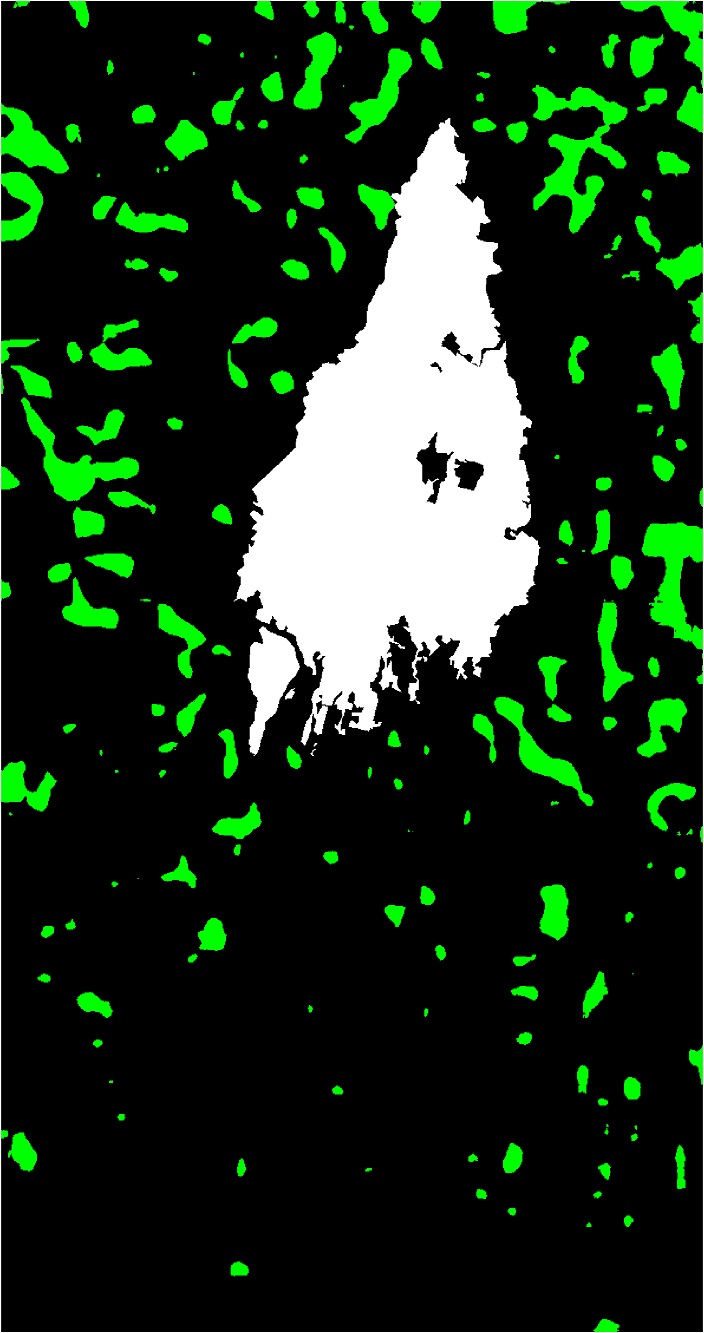}
\caption{\centering Training set,\newline $M = 10^5$, $k=20$}
\label{fig:ts20_tx}
\end{subfigure}
\end{center}
\caption{Affinity matrices comparison method for training set selection: training sets for $M=10^5$ and for $k=5$ (a) and $20$ (b).}
\label{fig:ts_tx}
\end{figure}

The comparison of histograms of the full dataset and selected training samples is illustrated by Fig.\ \ref{fig:hist_tx}, which shows an example for a two-channel subset of $\boldsymbol{Y}$.
$\mathcal{H}_{\boldsymbol{Y}}$ is depicted in blue, while $\mathcal{H}_{\boldsymbol{Y} \cap \mathcal{T}}$ is red for $k=20$ and for $M = 10^2,10^3,10^4,10^5$ in Fig.\ \ref{fig:hist_tx_1}, \ref{fig:hist_tx_2}, \ref{fig:hist_tx_3} and \ref{fig:hist_tx_4} respectively.
As commented previously, when $M$ increases, the histograms tend to present the same shape as $\mathcal{H}_{\boldsymbol{Y}}$.
However, they cover the domain of $\boldsymbol{Y}$ well even for $M = 10^2$, while avoiding both the outliers and the pixels belonging to changed areas.
In Fig.\ \ref{fig:ts5_tx} and \ref{fig:ts20_tx}, the pixels in the training set produced with $M = 10^5$ are depicted in green or red if they fall outside or inside the ground truth change area depicted in white, respectively.
It can be noticed that red points are very few or none.
All these results confirm a good overlap between the distributions of unchanged data and the training sets and, therefore, encourage us to proceed with the transformation.

\subsubsection{Image regression}

The hyperparameters of the GPR model are set after one iteration of a gradient ascent-based optimiser. 
For the SVR,  $\lambda=1$, $\epsilon=0.1$, and $\sigma=1$ are set following \cite{tuia2011multioutput}.
Concerning the RFR, $T = 64$, $r=[\nicefrac{P}{3}]$, $S = 1$ are chosen after a coarse grid search.
Lastly, the HPT is tuned by setting $K = 50$ and $\gamma = 100$, as empirically found in \cite{liu2018change}.

\begin{table*}[ht!]
\setlength{\tabcolsep}{4pt}
\small
\centering
\caption{Elapsed time in seconds during the regression, AUC after filtering, and OA and KC after thresholding for the regression methods applied on the Texas dataset with all considered combinations of $k$ and $M$. Best results in bold.}
\label{tab:Results_Tx}

\begin{tabular}{ c c | c c c c | c c c c | c c c c}
\toprule
\multicolumn{2}{c|}{Patch size $p$} & \multicolumn{4}{c|}{$k = 5$} & \multicolumn{4}{c|}{$k = 10$} & \multicolumn{4}{c}{$k = 20$} \\
\multicolumn{2}{c|}{Regression method} & GPR & SVR & RFR & HPT & GPR & SVR & RFR & HPT & GPR & SVR & RFR & HPT \\
\midrule
Elapsed & $M = 10^2$ & 11 &   1 &   6 & 126 &  11 &   1 &   6 & 110 &  11 &   1 &   6 & 104 \\
time & $M = 10^3$ & 90 &  11 &   6 & 131 &  92 &  11 &   6 & 126 &  90 &  11 &   6 & 125 \\
in & $M = 10^4$ & 914 & 107 &   8 & 178 & 912 & 107 &   8 & 180 & 913 & 106 &   8 & 179 \\
seconds & $M = 10^5$ & - &   - &  18 & 293 &   - &   - &  18 & 299 &   - &   - &  18 & 304 \\
\hline
\multirow{4}{*}{AUC} & $M = 10^2$ & 0.850 & 0.919 & 0.943 & 0.937 & 0.834 & 0.915 & 0.960 & 0.967 & 0.925 & 0.871 & 0.941 & 0.930 \\
& $M = 10^3$ & 0.940 & 0.969 & 0.965 & 0.972 & 0.914 & 0.962 & 0.952 & 0.957 & 0.921 & 0.971 & 0.971 & 0.973 \\
& $M = 10^4$ & 0.964 & \textbf{0.973} & 0.967 & 0.972 & 0.969 & 0.974 & 0.973 & 0.975 & 0.962 & 0.974 & 0.971 & 0.973 \\
& $M = 10^5$ & - & - & 0.959 & 0.966 & - & - & 0.974 & \textbf{0.976} & - & - & 0.976 & \textbf{0.977} \\
\hline
\multirow{4}{*}{OA} & $M = 10^2$ & 0.916 & 0.958 & 0.965 & 0.966 & 0.876 & 0.959 & 0.972 & 0.977 & 0.955 & 0.935 & 0.962 & 0.958 \\
& $M = 10^3$ & 0.959 & 0.981 & 0.979 & 0.982 & 0.926 & 0.979 & 0.965 & 0.968 & 0.929 & 0.982 & 0.982 & 0.983 \\
& $M = 10^4$ & 0.953 & 0.982 & 0.980 & \textbf{0.983} & 0.956 & 0.983 & 0.983 & \textbf{0.984} & 0.941 & 0.983 & 0.981 & 0.982 \\
& $M = 10^5$ & - & - & 0.975 & 0.980 & - & - & 0.982 & 0.983 & - & - & 0.983 & \textbf{0.984} \\
\hline
\multirow{4}{*}{KC} & $M = 10^2$ & 0.579 & 0.777 & 0.819 & 0.821 & 0.516 & 0.806 & 0.858 & 0.884 & 0.791 & 0.669 & 0.797 & 0.779 \\
& $M = 10^3$ & 0.791 & 0.901 & 0.892 & 0.906 & 0.620 & 0.893 & 0.811 & 0.832 & 0.627 & 0.903 & 0.904 & 0.908 \\
& $M = 10^4$ & 0.731 & 0.907 & 0.896 & \textbf{0.910} & 0.746 & 0.908 & 0.910 & \textbf{0.913} & 0.637 & 0.909 & 0.902 & 0.904 \\
& $M = 10^5$ & - & - & 0.865 & 0.895 & - & - & 0.905 & 0.908 & - & - & 0.909 & \textbf{0.914} \\
\bottomrule
\end{tabular}
\end{table*}

\begin{figure*}[ht!]
\begin{center}
\begin{subfigure}[t]{0.22\textwidth}
\includegraphics[width=\linewidth,keepaspectratio]{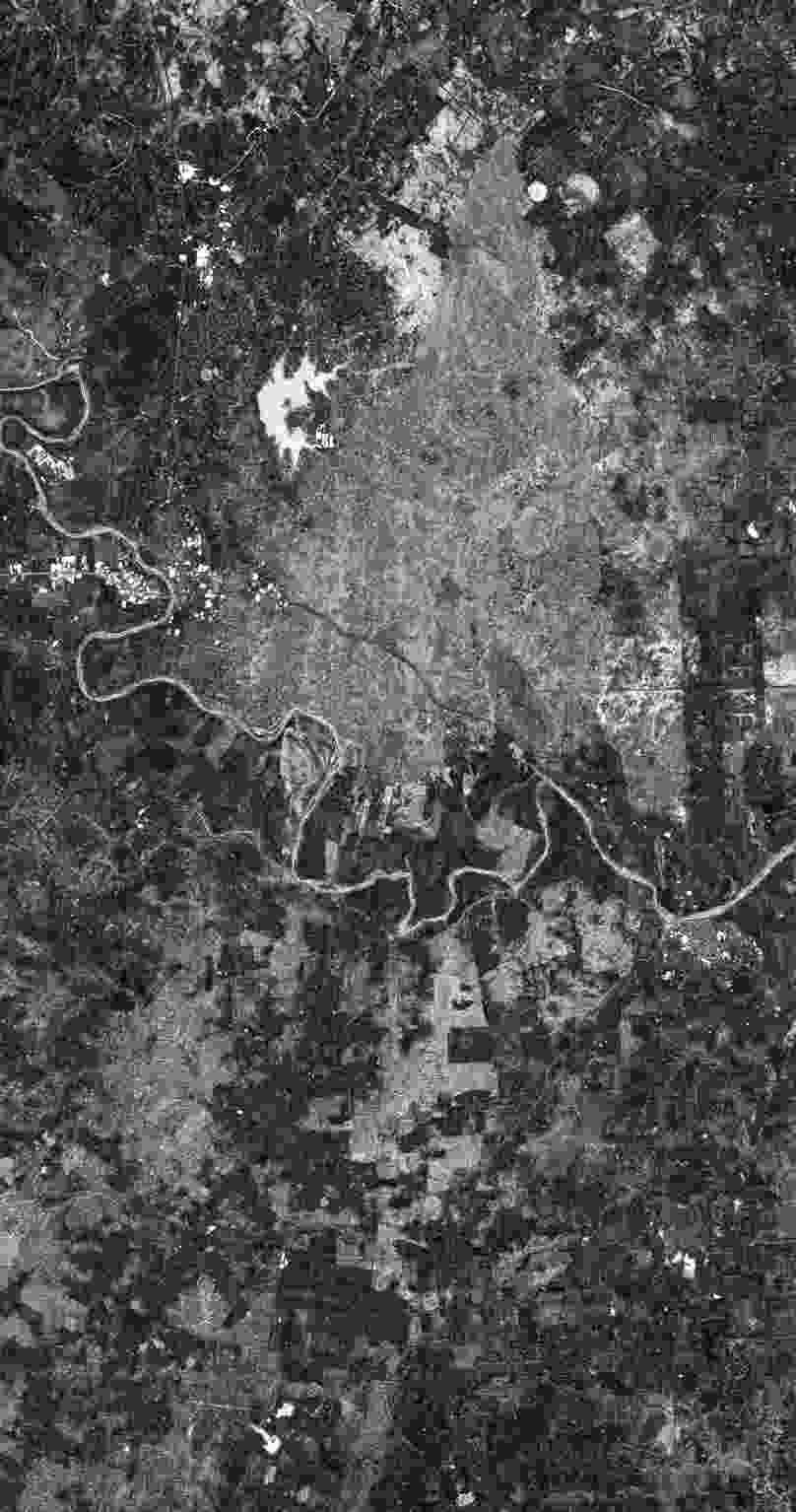}
\caption{\centering Worst result: GPR, \newline $M=10^2$, $k=10$}
\label{fig:subim1}
\end{subfigure}
\hspace{0.02\textwidth}%
\begin{subfigure}[t]{0.22\textwidth}
\includegraphics[width=\linewidth,keepaspectratio]{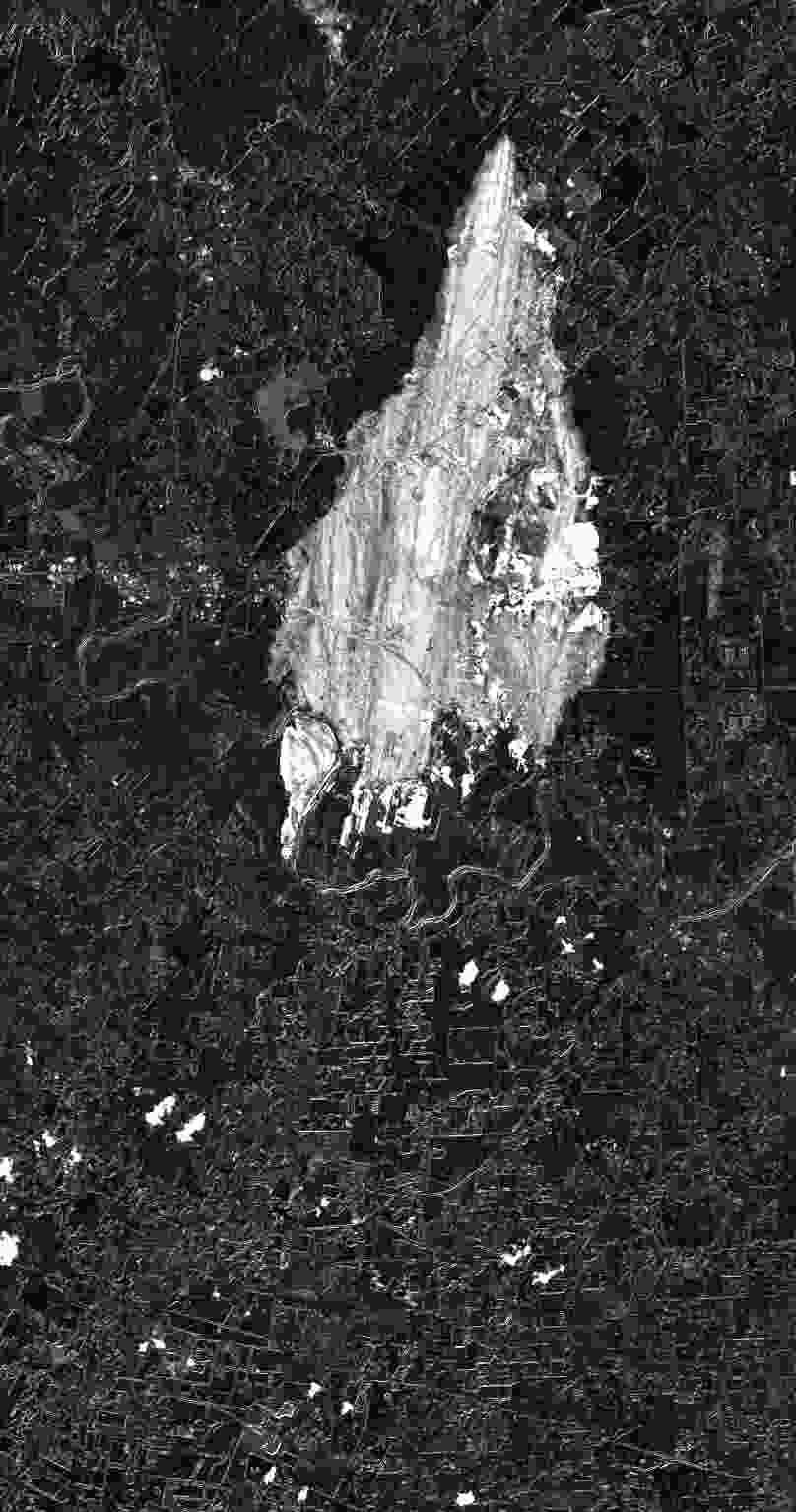}
\caption{\centering \hspace{0.5cm} SVR, \newline $M =10^4$, $k=20$}
\label{fig:subim2}
\end{subfigure}
\hspace{0.02\textwidth}%
\begin{subfigure}[t]{0.22\textwidth}
\includegraphics[width=\linewidth,keepaspectratio]{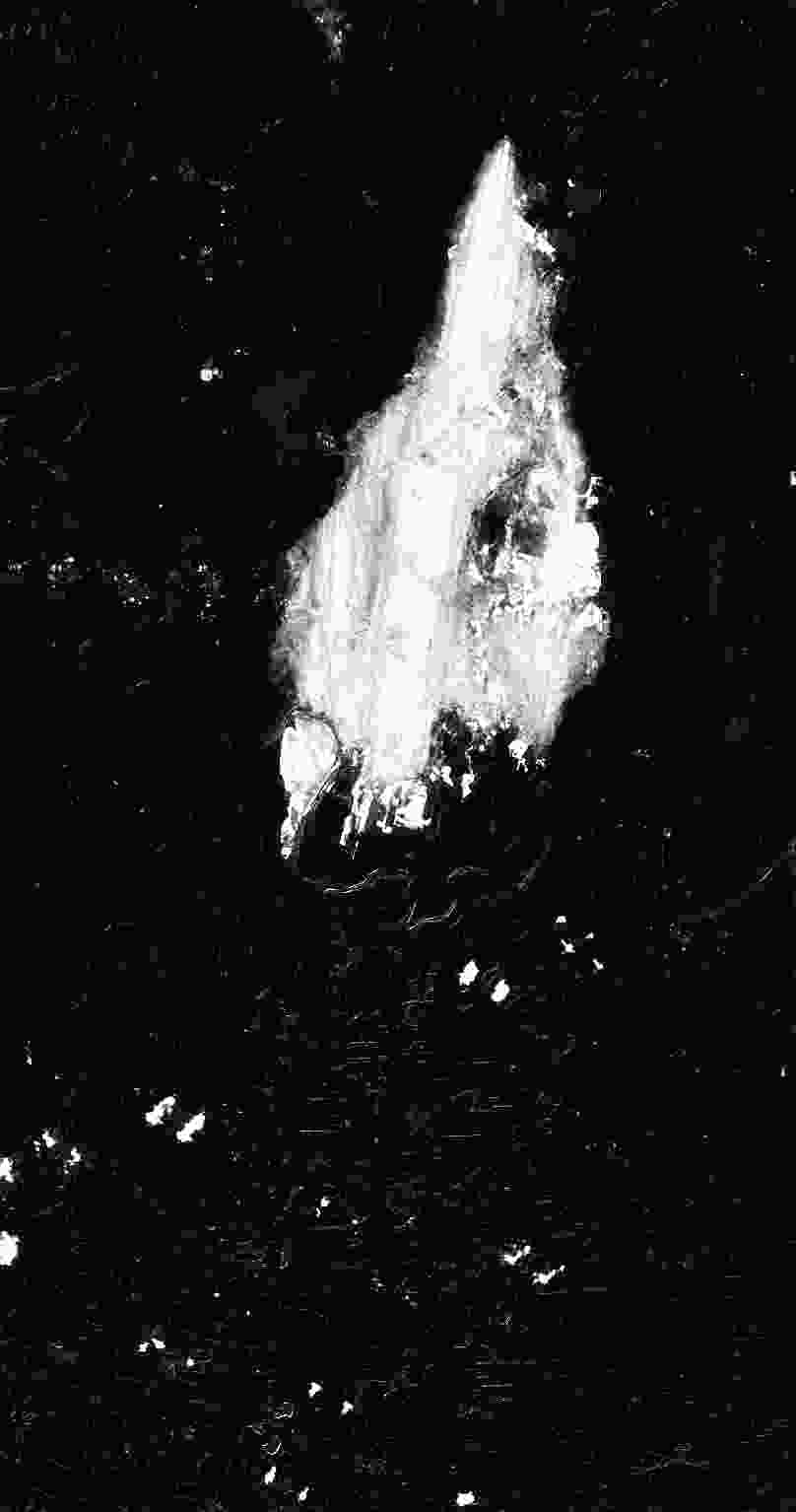}
\caption{\centering \hspace{0.2cm} SVR filtered, \newline $M =10^4$, $k=20$}
\label{fig:subim3}
\end{subfigure}
\hspace{0.02\textwidth}%
\begin{subfigure}[t]{0.22\textwidth}
\includegraphics[width=\linewidth,keepaspectratio]{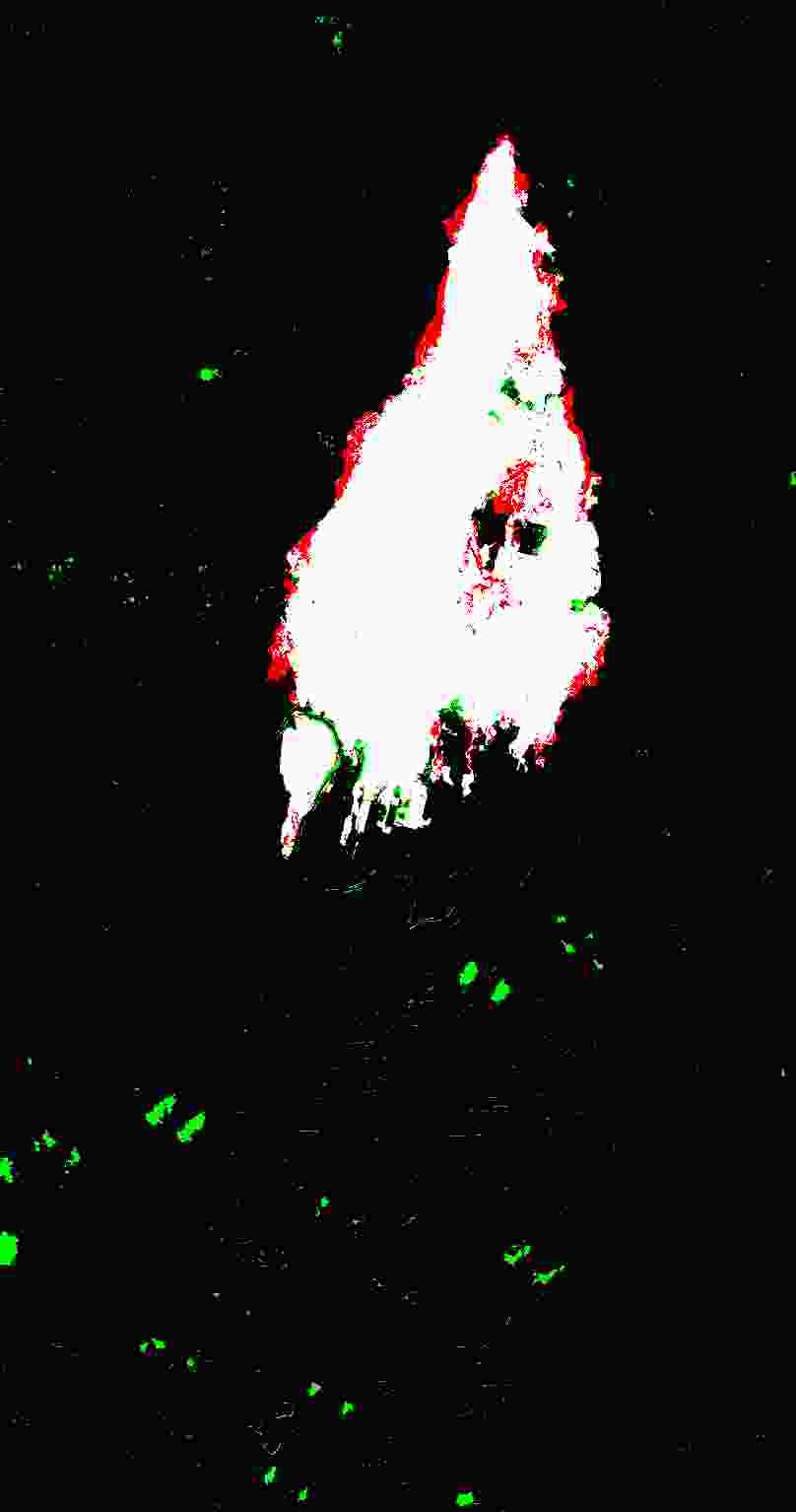}
\caption{\centering Confusion map, SVR, \newline $M=10^4$, $k=20$}
\label{fig:subim4}
\end{subfigure}
\end{center}

\caption{Some examples related to Table \ref{tab:Results_Tx}: (a) Result of the worst regression in all terms of accuracy produced by GPR with $M=10^2$ and $k=10$; (b) Example of successful regression produced by SVR with $M=10^4$ and $k=20$; (c) Corresponding output after filtering; (d) Corresponding confusion map after thresholding (c). White=TP, Green=FP, Red=FN, Black=TN.}
\label{fig:results1}
\end{figure*}

Table \ref{tab:Results_Tx} reports the results obtained for the proposed set of parameters and regression algorithms.
To put such outcomes in context, we report that the semi-supervised method proposed by Volpi \textit{et al}.\ in \cite{volpi2015spectral} could reach on average a KC of $0.65$ (standard deviation $0.06$).
The same methodology, improved by Yang \textit{et al}.\ in \cite{yang2018heterogeneous} by applying deep canonical correlation analysis, could not perform better than $0.947$ (standard deviation $0.02$) in OA and $0.71$ (standard deviation $0.1$) in KC.
In \cite{roscher2017sparse}, Roscher \textit{et al}.\ achieved a KC of $0.80$ on the same dataset but by using two Landsat 5 TM images, so performing homogeneous CD.
Overall, our unsupervised method consistently outperforms the aforementioned methods, both in terms of OA and KC.
We argue that the main reason behind these performance gaps is the size of the training sample.
These supervised methods rely on a training set which is hand-crafted, and it becomes unfeasible or unpractical to select more than few hundreds of data points.
Instead, our automatic selection provides a training sample which is one or two orders of magnitude larger.

As one may notice, the results for the SVR and the GPR are missing for $M=10^5$.
The computational time grows exponentially as $M$ increases, and for $M=10^5$ the size of the kernel matrices becomes too big, making the matrix multiplications unfeasible due to out of memory errors.
For brevity, only few examples of outcomes are depicted in Fig.\ \ref{fig:results1}.
Overall, every regression method yields good results, except for the GPR, which produced a rather low KC across all the examples.
For $M=10^2$, the regression methods are able to detect the changes, but there are many false positives due to the lake and the river, visible in Fig.\ \ref{fig:L5} and Fig.\ \ref{fig:ALI} and erroneously detected as change in Fig.\ \ref{fig:subim1}.
This means that for such a small $M$, the training set $\mathcal{T}$ does not include sufficient data samples from these classes.
The false positives shown as pairs of small, adjacent segments in Fig.\ \ref{fig:subim2}, Fig.\ \ref{fig:subim3} and Fig.\ \ref{fig:subim4} are clouds and their projected shadows, as it can be seen in Fig.\ \ref{fig:L5}.
The accuracy achieved by each regression approach on this dataset is consistently high.
On one hand, this image pair is not especially challenging in terms of heterogeneity.
On the other hand, it would be unfeasible to apply conventional CD methods designed for homogeneous data in this case.
Hence, this result demonstrates the effectiveness of the proposed regression-based approach to heterogeneous CD.

\subsection{Flood in California}

\begin{figure}[ht!]

\hspace{0.024\columnwidth}%
\begin{subfigure}[t]{0.3\columnwidth}
\includegraphics[width=\linewidth,keepaspectratio]{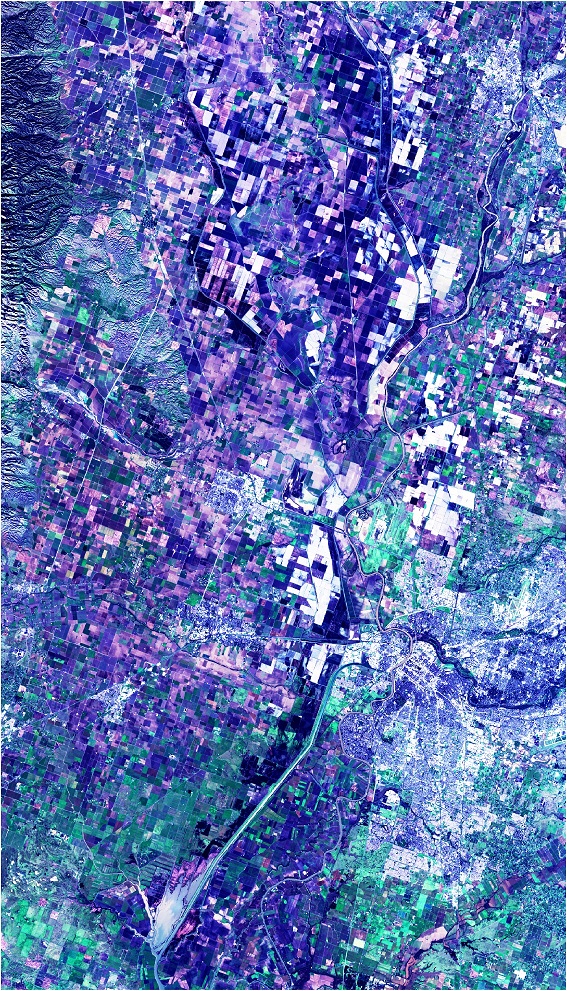} 
\caption{Landsat 8 ($t_1$)}
\label{fig2:L8}
\end{subfigure}
\hspace{0.024\columnwidth}%
\begin{subfigure}[t]{0.3\columnwidth}
\includegraphics[width=\linewidth,keepaspectratio]{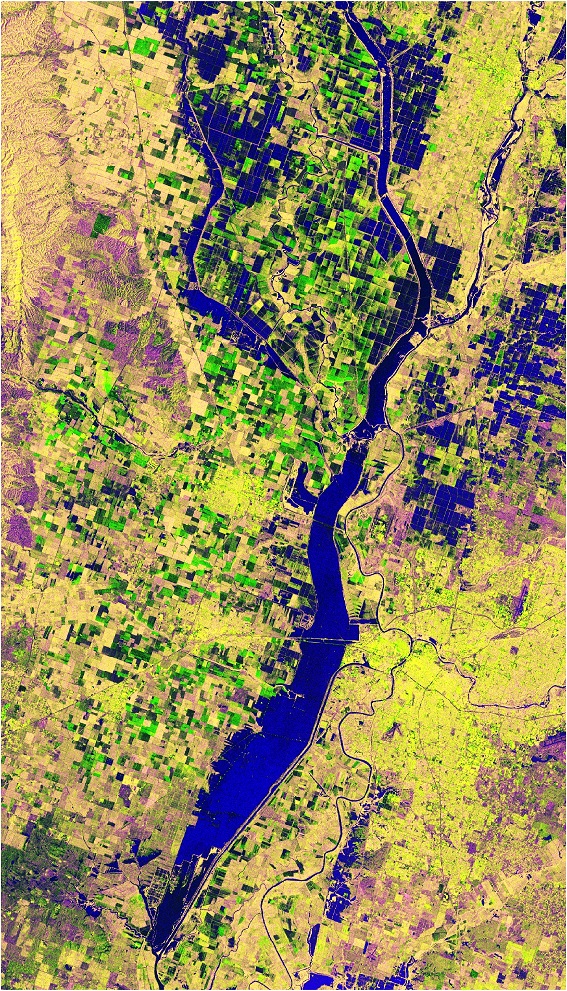}
\caption{Sentinel-1A ($t_2$)}
\label{fig2:S1A}
\end{subfigure}
\hspace{0.024\columnwidth}%
\begin{subfigure}[t]{0.3\columnwidth}
\includegraphics[width=\linewidth,keepaspectratio]{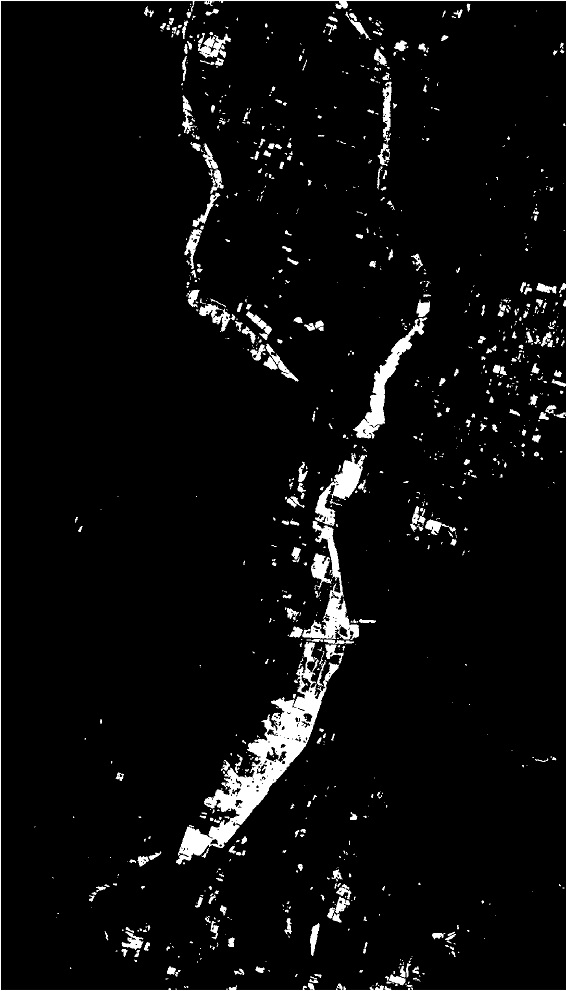}
\caption{Ground Truth}
\label{fig2:GT}
\end{subfigure}
\hspace{0.024\columnwidth}%

\caption{Flood in California: (a) Landsat 8 ($t_1$), (b) Sentinel-1A ($t_2$), (c) ground truth.}
\label{fig2:dataset2}
\end{figure}

\begin{figure}[b!]
\begin{center}
\begin{subfigure}[t]{0.4\columnwidth}
\includegraphics[width=\linewidth,keepaspectratio]{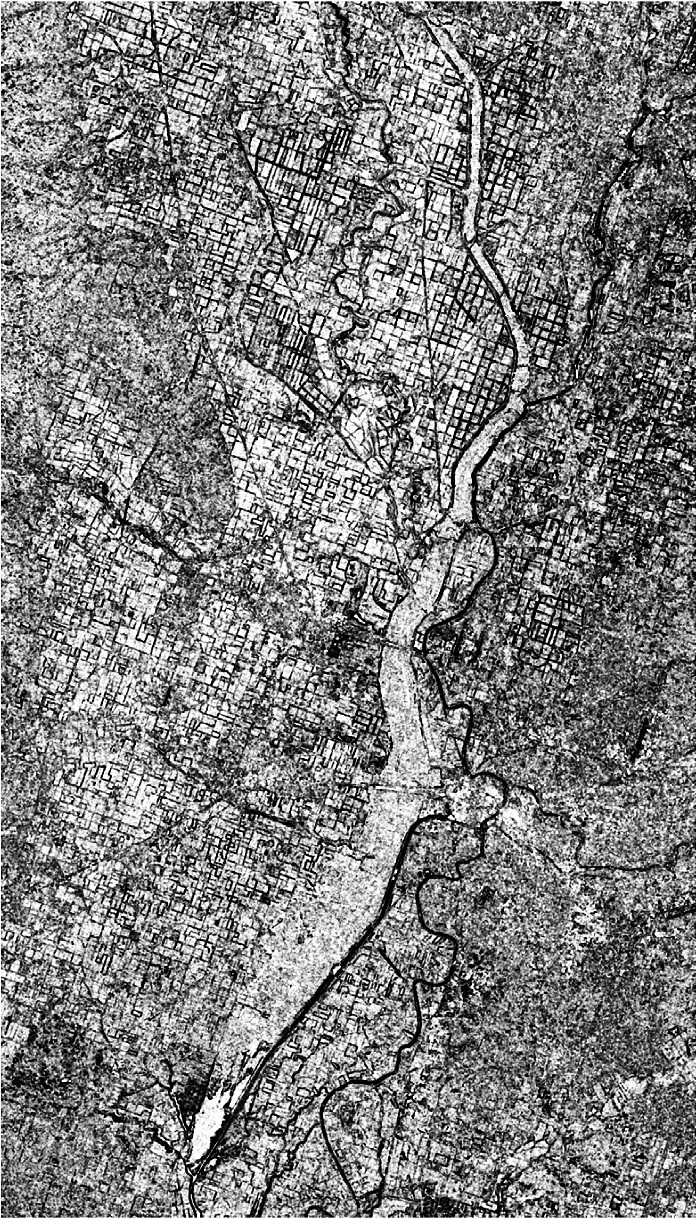} 
\caption{Heat-map, $k=5$}
\label{fig:hm5_cal}
\end{subfigure}
\hspace{0.05\columnwidth}%
\begin{subfigure}[t]{0.4\columnwidth}
\includegraphics[width=\linewidth,keepaspectratio]{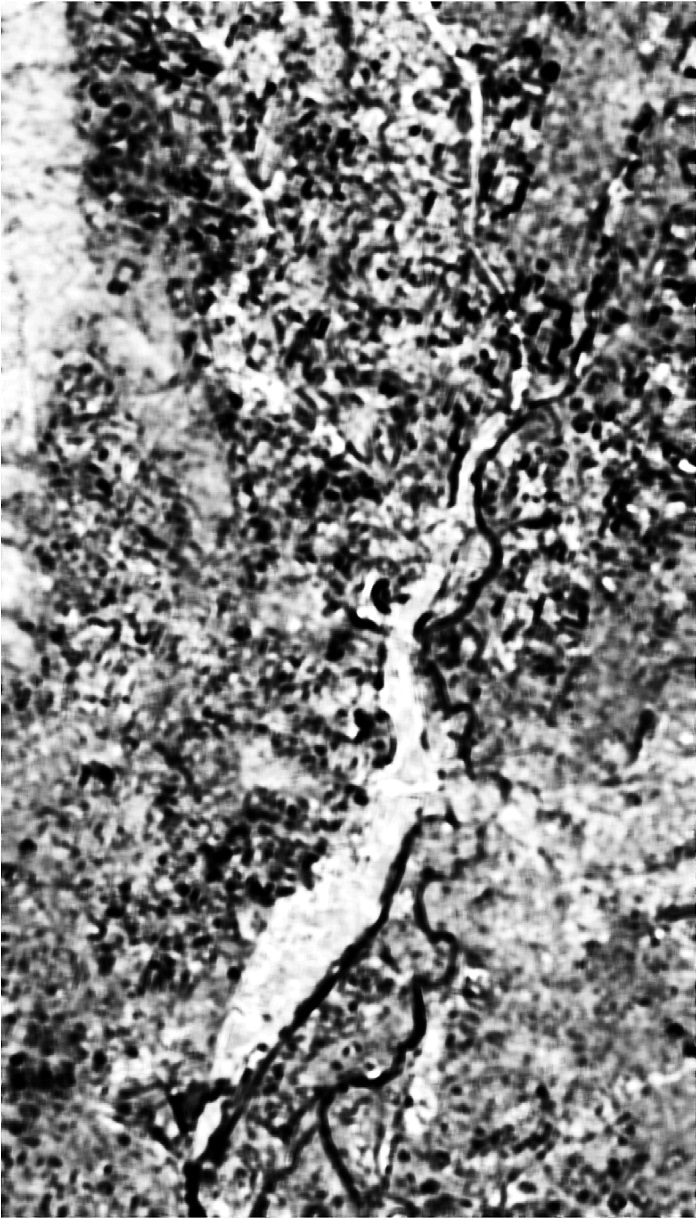}
\caption{Heat-map, $k=20$}
\label{fig:hm20_cal}
\end{subfigure}
\end{center}
\caption{Affinity matrices comparison method for training set selection: Heat-maps for $k=5$ (a) and $k=20$ (b).}
\label{fig:affm_cal}
\end{figure}

The second dataset represents a more challenging scenario, as it involves an multispectral image and a SAR image.
The image at time $t_1$ is a Landsat 8 acquisition\footnote{Distributed by LP DAAC, http://lpdaac.usgs.gov} covering Sacramento County, Yuba County and Sutter County, California, on 5 January 2017.
It is composed of nine channels covering the spectrum from deep blue to short-wave infrared, plus two long-wave infrared channels (Fig.\ \ref{fig2:L8} shows the RGB channels).
The time $t_2$ image was acquired on 18 February 2017 by Sentinel-1A\footnote{Data processed by ESA, http://www.copernicus.eu/} over the same area after the occurrence of a flood.
The image is recorded in polarisations VV and VH, and the dataset is augmented with the ratio between the two intensities as a third channel, yielding the false colour RGB image shown in Fig.\ \ref{fig2:S1A}.
To obtain a reasonable ground truth without resorting to manual selection, we used two other single-polarisation SAR images acquired approximately at the same times as the previous ones.
The ground truth is extracted from these two images and is shown in Fig.\ \ref{fig2:GT}.

\subsubsection{Training set selection}

\begin{figure}[ht!]
\begin{center}
\begin{subfigure}[t]{0.49\columnwidth}
\includegraphics[width=\linewidth,keepaspectratio]{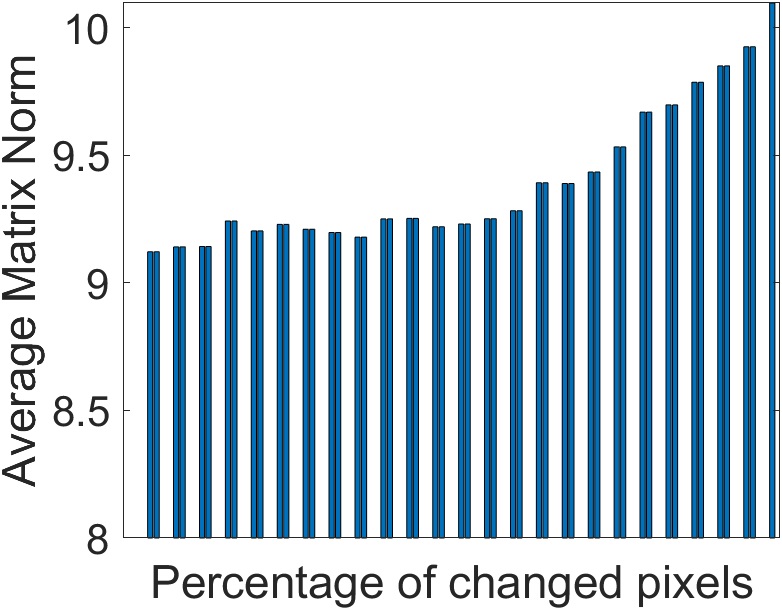}
\caption{$k=5$}
\label{fig:mean_f_cal_5}
\end{subfigure}~%
\begin{subfigure}[t]{0.49\columnwidth}
\includegraphics[width=\linewidth,keepaspectratio]{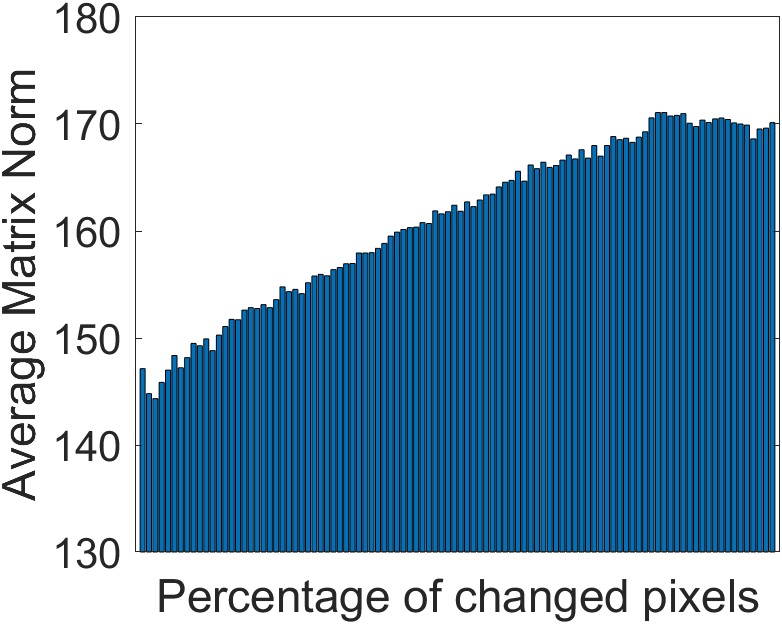}
\caption{$k=20$}
\label{fig:mean_f_cal_20}
\end{subfigure}
\end{center}
\caption{Average matrix norm $f$ versus the percentage of changed pixels for $k=5$ (a) and $k=20$ (b).}
\label{fig:mean_f_cal}
\end{figure}

As in the previous case, the heat-maps from the self-supervised training set selection with $k=5$ and $k=20$ are depicted in Fig.\ \ref{fig:affm_cal}.
For change detection directly on the heat-maps we achieve an AUC equal to $64.9\%$, $71.6\%$, and $76.0\%$ in about $20$ minutes, $2$ hours and $5$ hours for $k=5,10,20$ respectively.
Unlike in the experiment with the Texas dataset, the $5 \times 5$ and the $10 \times 10$ sliding windows do not provide a preliminary change feature with the same quality in this case.
The average matrix norm $f$ with respect to the percentage of changed pixels is shown in Fig.\ \ref{fig:mean_f_cal}, for $k=5$ and $k=20$.
In this case the trend is not so obvious as before, especially for $k=5$, but it is still recognisable.

\begin{figure*}[b!]

\begin{center}
\begin{subfigure}[t]{0.49\columnwidth}
\includegraphics[width=\linewidth,keepaspectratio]{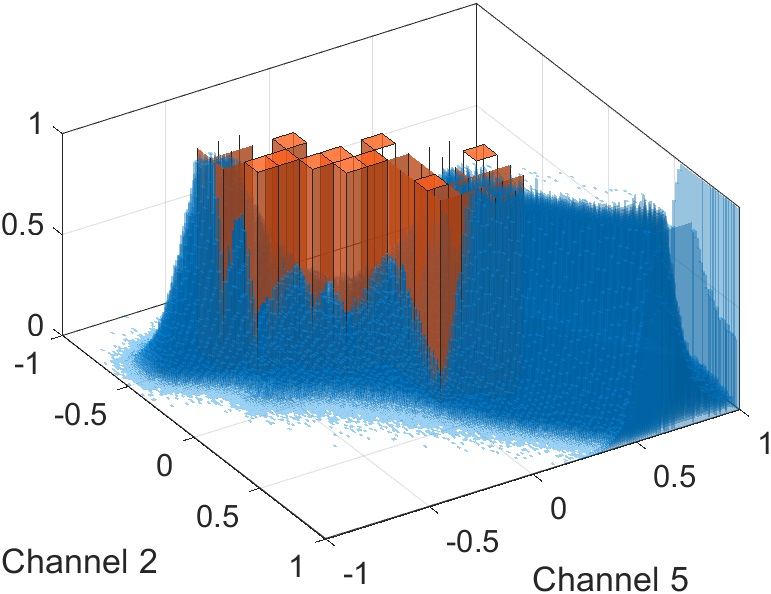}
\caption{\centering $\mathcal{H}_{\boldsymbol{Y}}$ (blue), $\mathcal{H}_{\boldsymbol{Y} \cap \mathcal{T}}$ (red) \newline $M=10^2$}
\label{fig:hist_cal_1}
\end{subfigure}
\begin{subfigure}[t]{0.49\columnwidth}
\includegraphics[width=\linewidth,keepaspectratio]{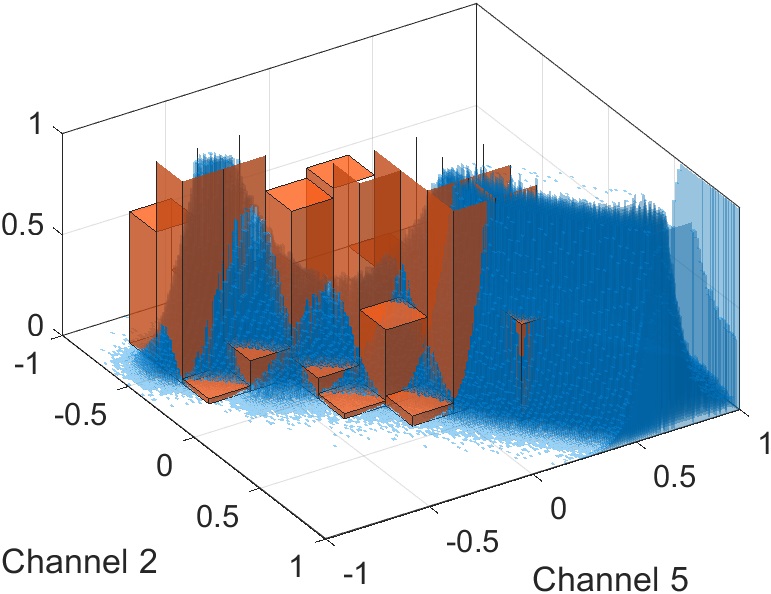}
\caption{\centering $\mathcal{H}_{\boldsymbol{Y}}$ (blue), $\mathcal{H}_{\boldsymbol{Y} \cap \mathcal{T}}$ (red) \newline $M=10^3$}
\label{fig:hist_cal_2}
\end{subfigure}
\begin{subfigure}[t]{0.49\columnwidth}
\includegraphics[width=\linewidth,keepaspectratio]{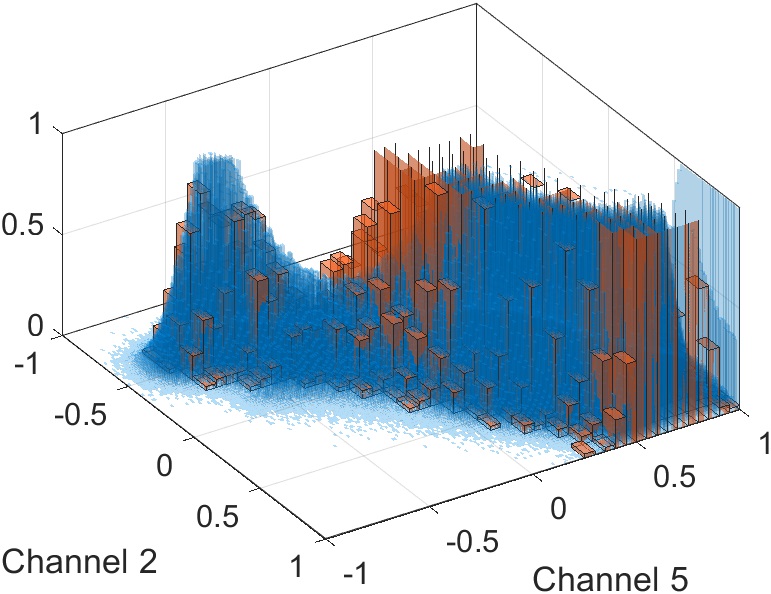}
\caption{\centering $\mathcal{H}_{\boldsymbol{Y}}$ (blue), $\mathcal{H}_{\boldsymbol{Y} \cap \mathcal{T}}$ (red) \newline $M=10^4$}
\label{fig:hist_cal_3}
\end{subfigure}
\begin{subfigure}[t]{0.49\columnwidth}
\includegraphics[width=\linewidth,keepaspectratio]{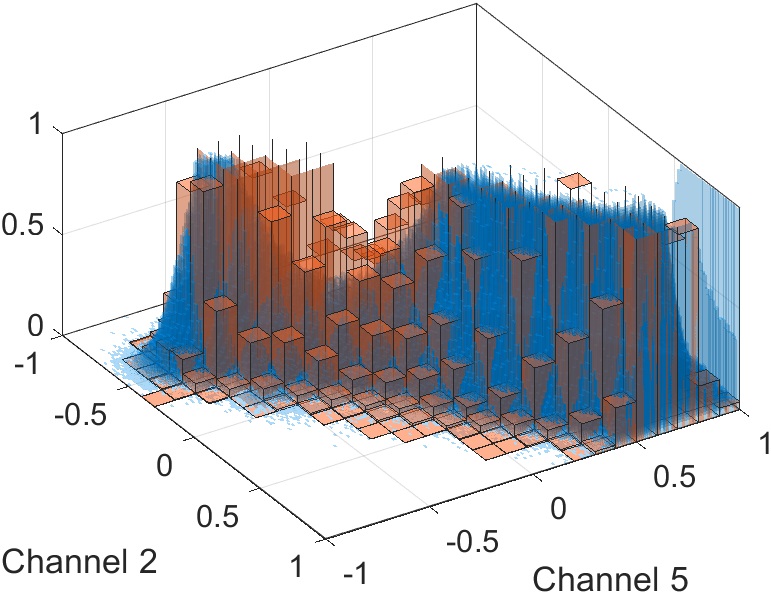}
\caption{\centering $\mathcal{H}_{\boldsymbol{Y}}$ (blue), $\mathcal{H}_{\boldsymbol{Y} \cap \mathcal{T}}$ (red) \newline $M=10^5$}
\label{fig:hist_cal_4}
\end{subfigure}
\end{center}

\caption{Example of comparison between $\mathcal{H}_{\boldsymbol{Y}}$ and $\mathcal{H}_{\boldsymbol{Y} \cap \mathcal{T}}$ with two channels of $\boldsymbol{Y}$ for $k=20$: $\mathcal{H}_{\boldsymbol{Y}}$ is depicted in blue, whereas $\mathcal{H}_{\boldsymbol{Y} \cap \mathcal{T}}$ is depicted in red for $M = 10^2,10^3,10^4,10^5$ in (a), (b), (c) and (d) respectively. Best viewed in colour.}

\label{fig:hist_cal}
\end{figure*}

\begin{table}[ht!]
\setlength{\tabcolsep}{4pt}
\small
\centering
\caption{AUC, FN, $d_H\left(\mathcal{H}_{\boldsymbol{X}},\mathcal{H}_{\boldsymbol{X} \cap \mathcal{T}}\right)$ and $d_H\left(\mathcal{H}_{\boldsymbol{Y}},\mathcal{H}_{\boldsymbol{Y} \cap \mathcal{T}}\right)$ for various combinations of $k$ and $M$ on the California dataset.}
\label{tab:Aff_cal}

\begin{tabular}{ c c | c c c}
\toprule
\multicolumn{2}{c|}{Patch size $k$} & $k = 5$ & $k = 10$ & $k = 20$ \\
\midrule
 \multicolumn{2}{c|}{AUC} & 64.9 & 71.6 & 76.0\\
 \hline
\multirow{4}{*}{FN} & $M = 10^2$ & 2.000 & 0 & 0 \\
& $M = 10^3$ & 2.400 & 4.000 & 0 \\
& $M = 10^4$ & 1.970 & 2.350 & 0.560 \\
& $M = 10^5$ & 1.961 & 2.048 & 0.831 \\
\hline
\multirow{4}{*}{$d_H\left(\mathcal{H}_{\boldsymbol{X}},\mathcal{H}_{\boldsymbol{X} \cap \mathcal{T}}\right)$} & $M = 10^2$ & 0.768 & 0.781 & 0.755 \\
& $M = 10^3$ & 0.447 & 0.553 & 0.466 \\
& $M = 10^4$ & 0.308 & 0.339 & 0.291 \\
& $M = 10^5$ & 0.209 & 0.202 & 0.194 \\
\hline
\multirow{4}{*}{$d_H\left(\mathcal{H}_{\boldsymbol{Y}},\mathcal{H}_{\boldsymbol{Y} \cap \mathcal{T}}\right)$} & $M = 10^2$ & 0.801 & 0.796 & 0.800 \\
& $M = 10^3$ & 0.498 & 0.557 & 0.439 \\
& $M = 10^4$ & 0.332 & 0.335 & 0.229 \\
& $M = 10^5$ & 0.230 & 0.194 & 0.144 \\
\bottomrule
\end{tabular}
\end{table}

Along with the AUC, Table \ref{tab:Aff_cal} reports the percentage of changed pixels included in $\mathcal{T}$, denoted as FN, and the Hellinger distances for the considered values of $k$ and $M$. The Hellinger distances 
$d_H\left(\mathcal{H}_{\boldsymbol{X}},\mathcal{H}_{\boldsymbol{X} \cap \mathcal{T}}\right)$ and $d_H\left(\mathcal{H}_{\boldsymbol{Y}},\mathcal{H}_{\boldsymbol{Y} \cap \mathcal{T}}\right)$ are very high for $M = 10^2$ and $M = 10^3$.
This does not come unexpected: Given the huge difference between $M$ and the total number of data points, which is $N = 7\cdot10^6$ for this dataset, it is impossible for the histogram of the training points $\mathcal{H}_{\boldsymbol{Y} \cap \mathcal{T}}$ to be similar to the histogram of the whole image $\mathcal{H}_{\boldsymbol{Y}}$.
Differently from the previous scenario, $d_H\left(\mathcal{H}_{\boldsymbol{X}},\mathcal{H}_{\boldsymbol{X} \cap \mathcal{T}}\right)$ and $d_H\left(\mathcal{H}_{\boldsymbol{Y}},\mathcal{H}_{\boldsymbol{Y} \cap \mathcal{T}}\right)$ are very similar and follow the same pattern in Table \ref{tab:Aff_cal}, decreasing as the number of training points grows and as the patch size $k$ becomes larger.
This is because the changes are not represented by a new class, as it was for the Texas forest fire.
Instead, the flooded areas have features very similar to, for example, the ones of the many rice fields spread across the covered counties.
However, the methods avoids to involve the changed pixels in $\mathcal{T}$, as the FN in Table \ref{tab:Aff_cal} show.
Fig.\ \ref{fig:hist_cal} provides an example for $\boldsymbol{Y}$ and its first two channels, where it is shown how $\mathcal{H}_{\boldsymbol{Y} \cap \mathcal{T}}$ first covers the domain of $\boldsymbol{Y}$ when $M = 10^2$, and then from $M = 10^3$ to $M = 10^5$ it gradually changes its shape and fits better to $\mathcal{H}_{\boldsymbol{Y}}$.

\begin{figure}[ht!]
\begin{center}
\begin{subfigure}[t]{0.4\columnwidth}
\includegraphics[width=\linewidth,keepaspectratio]{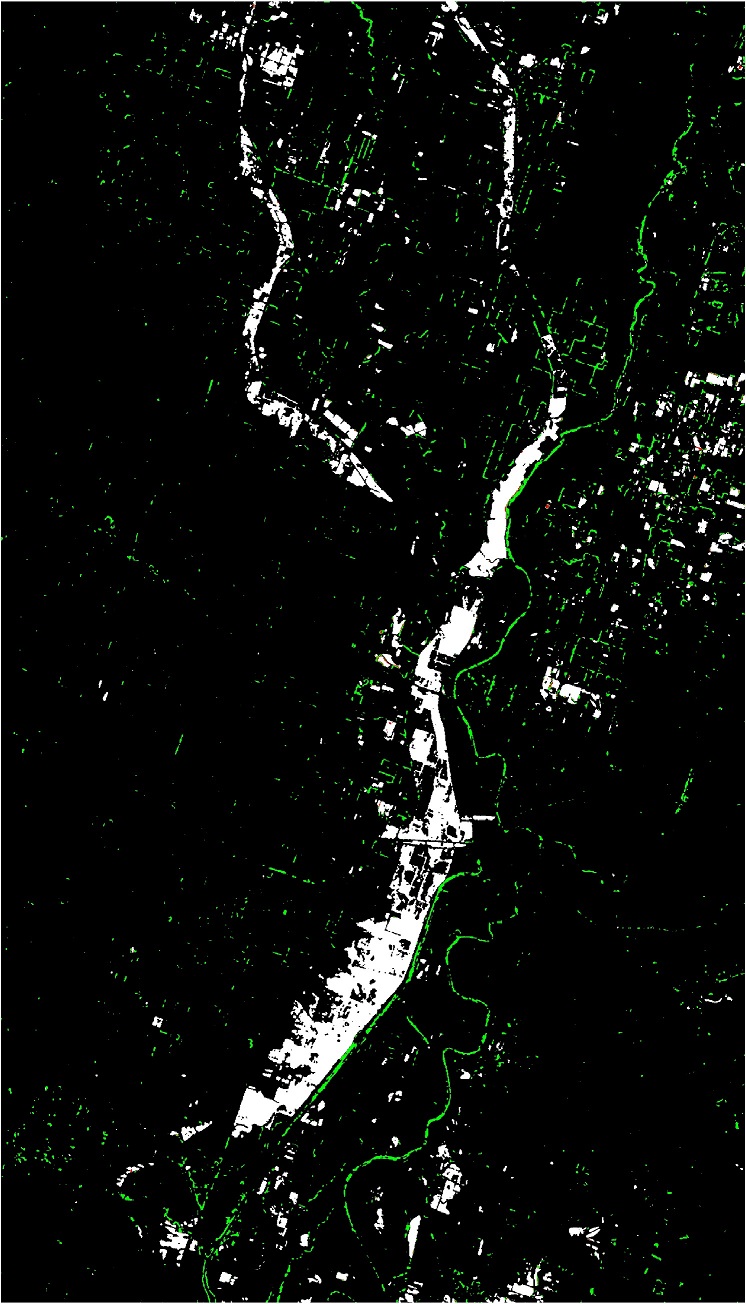}
\caption{\centering Training set,\newline $M = 10^5$, $k=5$}
\label{fig:ts5_cal}
\end{subfigure}
\hspace{0.05\columnwidth}%
\begin{subfigure}[t]{0.4\columnwidth}
\includegraphics[width=\linewidth,keepaspectratio]{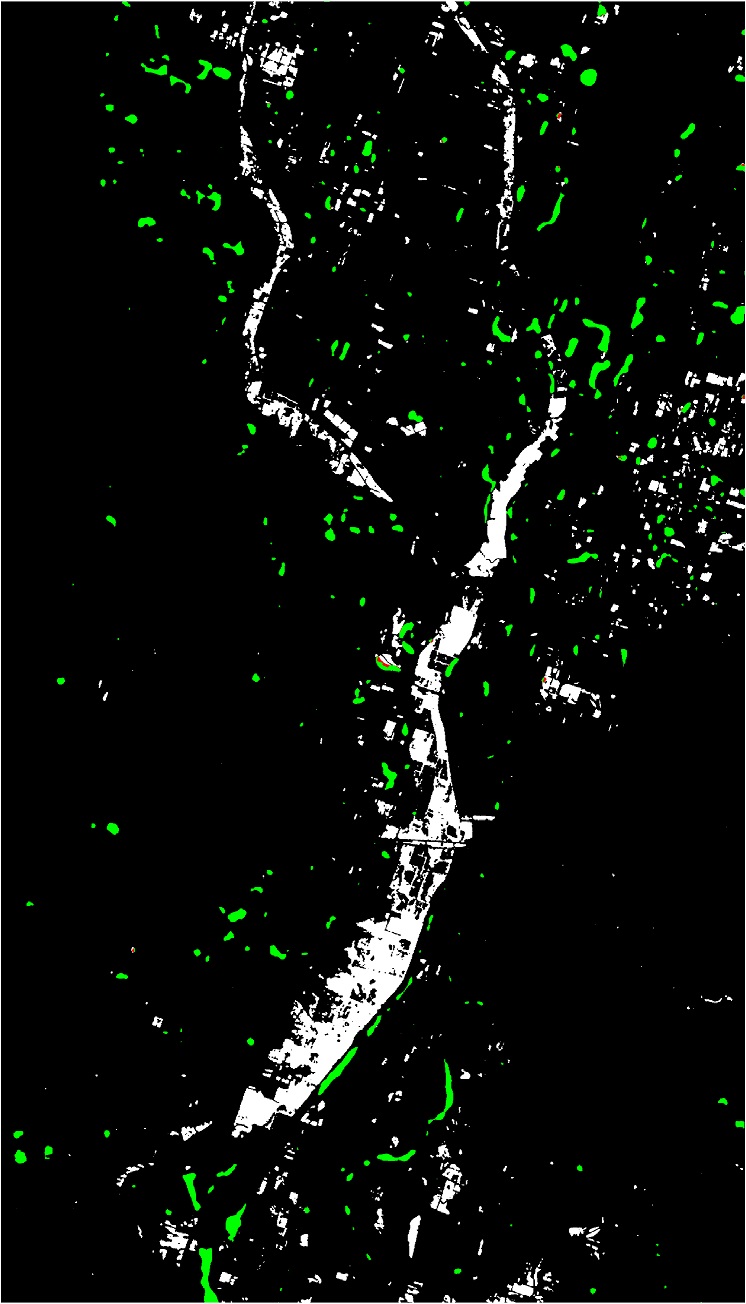}
\caption{\centering Training set,\newline $M = 10^5$, $k=20$}
\label{fig:ts20_cal}
\end{subfigure}
\end{center}
\caption{Affinity matrices comparison method for training set selection: selected training sets for $M=10^5$ and for $k=5$ (a) and $k=20$ (b).}
\label{fig:ts_cal}
\end{figure}

Fig.\ \ref{fig:ts_cal} illustrates two examples of training sets for $M = 10^5$, for $k=5$ in Fig.\ \ref{fig:ts5_tx} and $k=20$ in Fig.\ \ref{fig:ts20_tx}. The data points are depicted in green or red if they fall outside or inside the changed areas depicted in white, respectively (red points are very few).

\subsubsection{Image regression}

The four regression methods were tuned following the same procedure as before.

\begin{table*}[ht!]
\centering
\setlength{\tabcolsep}{4pt}
\small
\caption{Elapsed time in seconds after the regression, AUC after filtering, and OA and KC after thresholding, for the regression methods applied on the California dataset with all the considered combinations of $k$ and $M$. Best results in bold.}
\label{tab:Results_cal}
\begin{tabular}{ c c | c c c c | c c c c | c c c c}
\toprule
\multicolumn{2}{c|}{Patch size $p$} & \multicolumn{4}{c|}{$k = 5$} & \multicolumn{4}{c|}{$k = 10$} & \multicolumn{4}{c}{$k = 20$} \\
\multicolumn{2}{c|}{Regression method} & GPR & SVR & RFR & HPT & GPR & SVR & RFR & HPT & GPR & SVR & RFR & HPT \\
\midrule
Elapsed & $M = 10^2$ &  77 &   8 &  25 & 585 &  80 &   8 &  24 & 597 &  78 &   8 &  25 & 590 \\
time & $M = 10^3$ & 508 &  80 &  28 & 650 & 508 &  84 &  27 & 637 & 507 &  80 &  28 & 649 \\
in & $M = 10^4$ & 4796 & 699 &  35 & 775 & 4802 & 707 &  34 & 795 & 4793 & 699 &  35 & 775 \\
seconds & $M = 10^5$ & - &   - &  59 & 2015 &   - &   - &  60 & 2023 &   - &   - &  60 & 2008 \\
\hline
\multirow{4}{*}{AUC} & $M = 10^2$ & 0.790 & \textbf{0.889} & 0.850 & 0.833 & 0.835 & \textbf{0.881} & 0.880 & 0.875 & 0.830 & 0.860 & 0.853 & 0.854 \\
& $M = 10^3$ & 0.821 & 0.879 & 0.860 & 0.853 & 0.795 & 0.857 & 0.865 & 0.849 & 0.745 & 0.884 & 0.894 & 0.882 \\
& $M = 10^4$ & 0.688 & 0.860 & 0.847 & 0.849 & 0.728 & 0.870 & 0.867 & 0.862 & 0.784 & \textbf{0.905} & 0.901 & 0.897 \\
& $M = 10^5$ & - & - & 0.865 & 0.868 & - & - & 0.876 & 0.877 & - & - & 0.903 & 0.902 \\
\hline
\multirow{4}{*}{OA} & $M = 10^2$ & 0.895 & 0.919 & 0.888 & 0.886 & 0.907 & 0.888 & 0.881 & 0.894 & 0.890 & 0.911 & 0.892 & 0.903 \\
& $M = 10^3$ & 0.933 & 0.916 & 0.916 & 0.913 & 0.903 & 0.913 & 0.915 & 0.908 & 0.912 & 0.915 & 0.925 & 0.926 \\
& $M = 10^4$ & 0.895 & 0.915 & 0.910 & 0.909 & 0.870 & 0.921 & 0.921 & 0.918 & 0.875 & 0.930 & 0.929 & 0.927 \\
& $M = 10^5$ & - & - & \textbf{0.922} & 0.921 & - & - & \textbf{0.926} & 0.924 & - & - & \textbf{0.933} & 0.930 \\
\hline
\multirow{4}{*}{KC} & $M = 10^2$ & 0.220 & 0.418 & 0.320 & 0.311 & 0.224 & 0.328 & 0.316 & 0.340 & 0.227 & 0.363 & 0.317 & 0.340 \\
& $M = 10^3$ & 0.186 & 0.393 & 0.381 & 0.367 & 0.043 & 0.370 & 0.385 & 0.354 & 0.198 & 0.377 & 0.421 & 0.414 \\
& $M = 10^4$ & 0.089 & 0.371 & 0.347 & 0.338 & 0.061 & 0.396 & 0.399 & 0.383 & 0.208 & 0.446 & 0.442 & 0.431 \\
& $M = 10^5$ & - & - & \textbf{0.399} & 0.393 & - & - & \textbf{0.415} & 0.408 & - & - & \textbf{0.462} & 0.450 \\
\bottomrule
\end{tabular}
\end{table*}

\begin{figure*}[ht!]
\begin{center}
\begin{subfigure}[t]{0.22\textwidth}
\includegraphics[width=\linewidth,keepaspectratio]{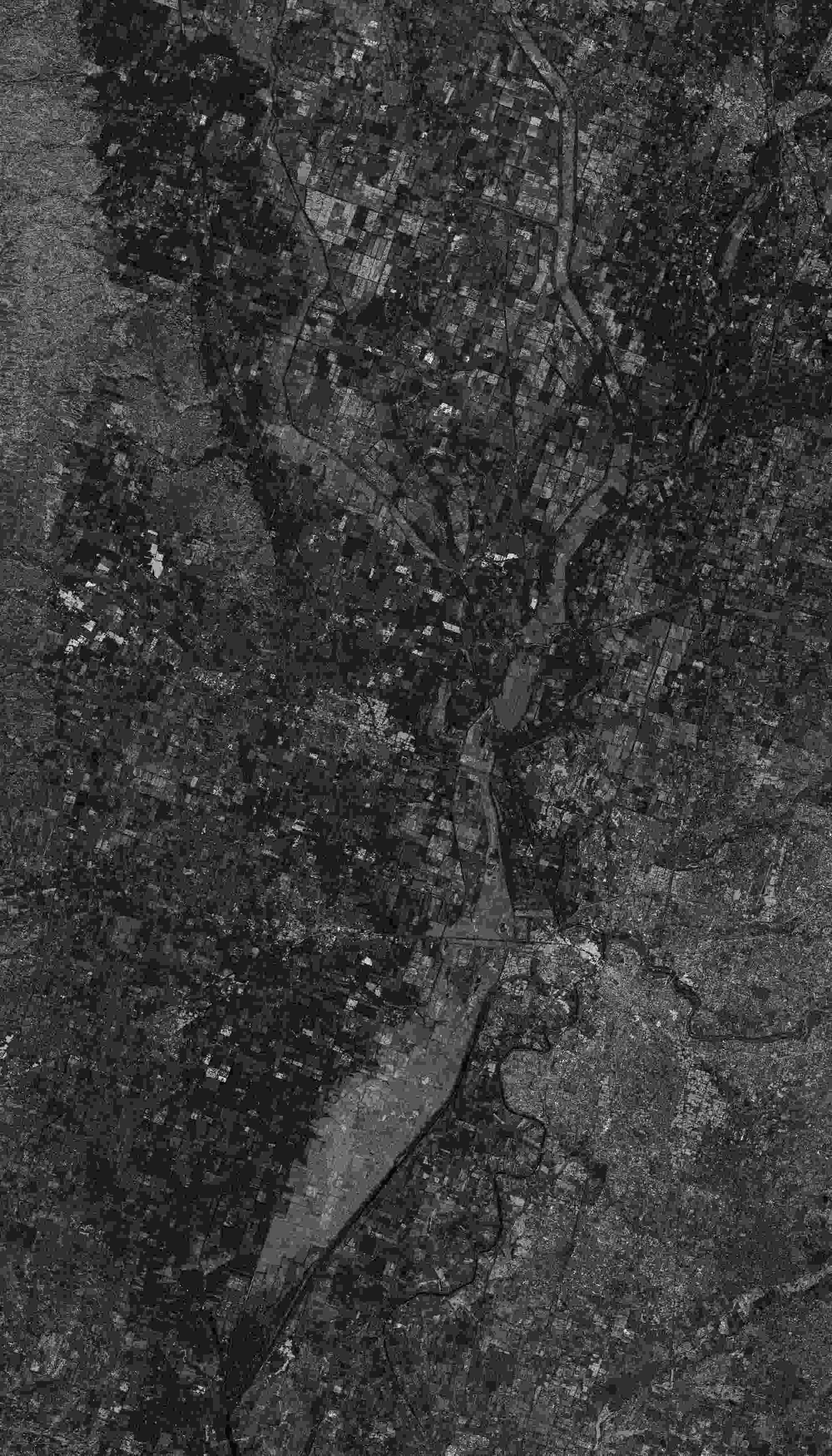}
\caption{\centering Worst result: GPR,\newline $M=10^4$, $k=10$}
\label{fig:GP10_10K_cal}
\end{subfigure}
\hspace{0.02\textwidth}%
\begin{subfigure}[t]{0.22\textwidth}
\includegraphics[width=\linewidth,keepaspectratio]{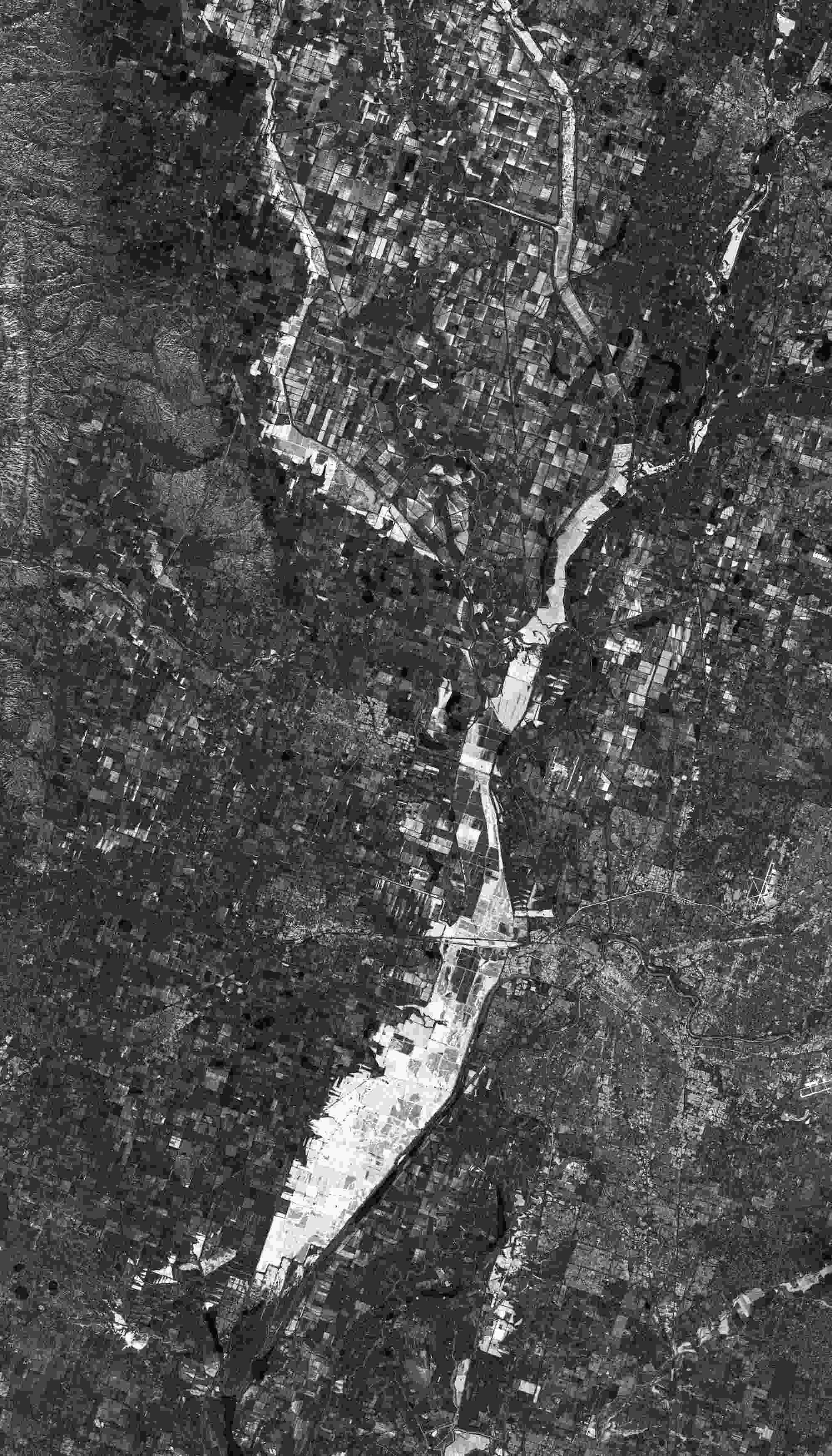}
\caption{\centering \hspace{0.5cm} RFR, \newline $M =10^5$, $k=20$}
\label{fig:RF20_100K_cal}
\end{subfigure}
\hspace{0.02\textwidth}%
\begin{subfigure}[t]{0.22\textwidth}
\includegraphics[width=\linewidth,keepaspectratio]{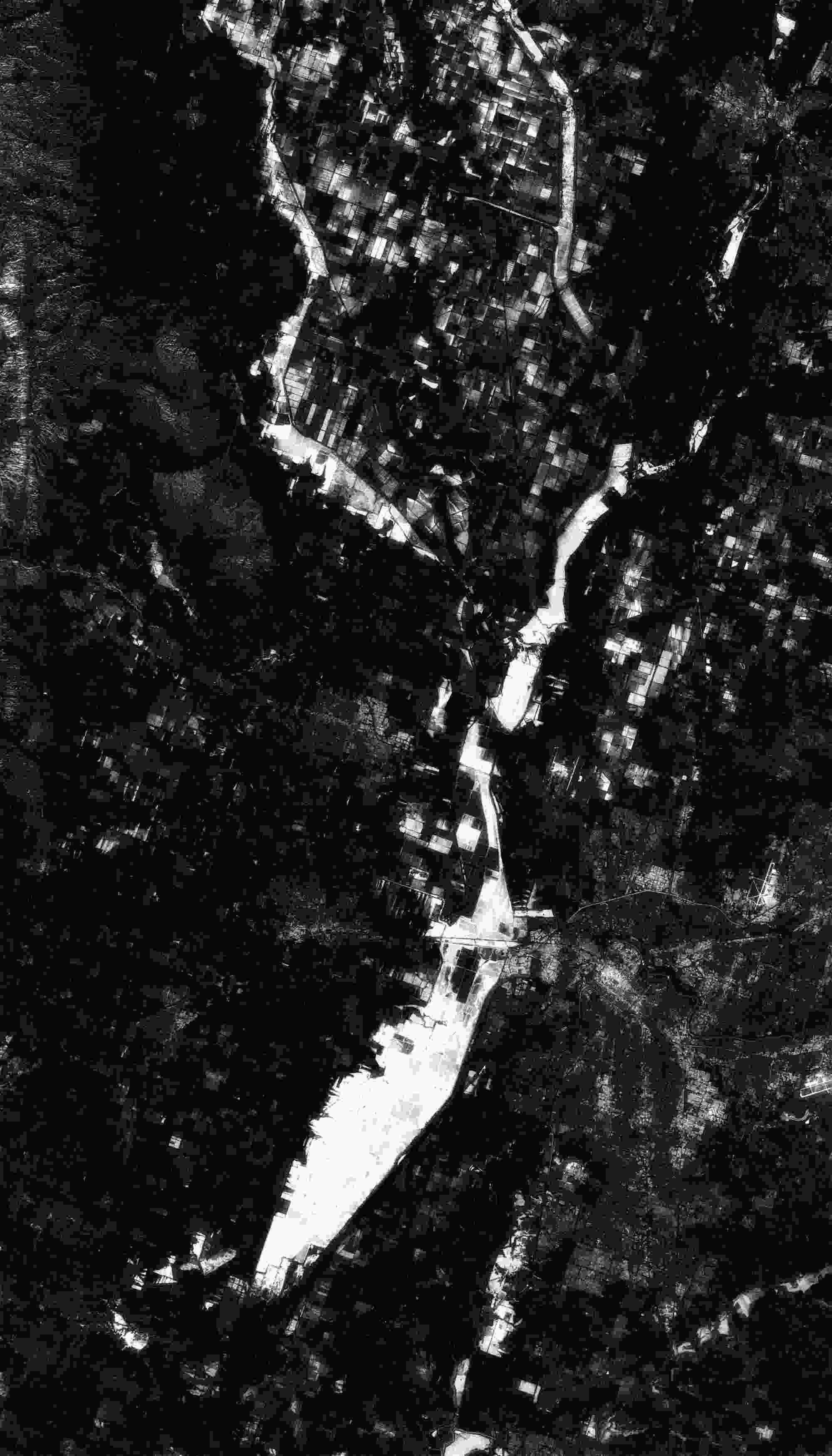}
\caption{\centering \hspace{0.2cm} RFR filtered, \newline $M =10^5$, $k=20$}
\label{fig:filtered_cal}
\end{subfigure}
\hspace{0.02\textwidth}%
\begin{subfigure}[t]{0.22\textwidth}
\includegraphics[width=\linewidth,keepaspectratio]{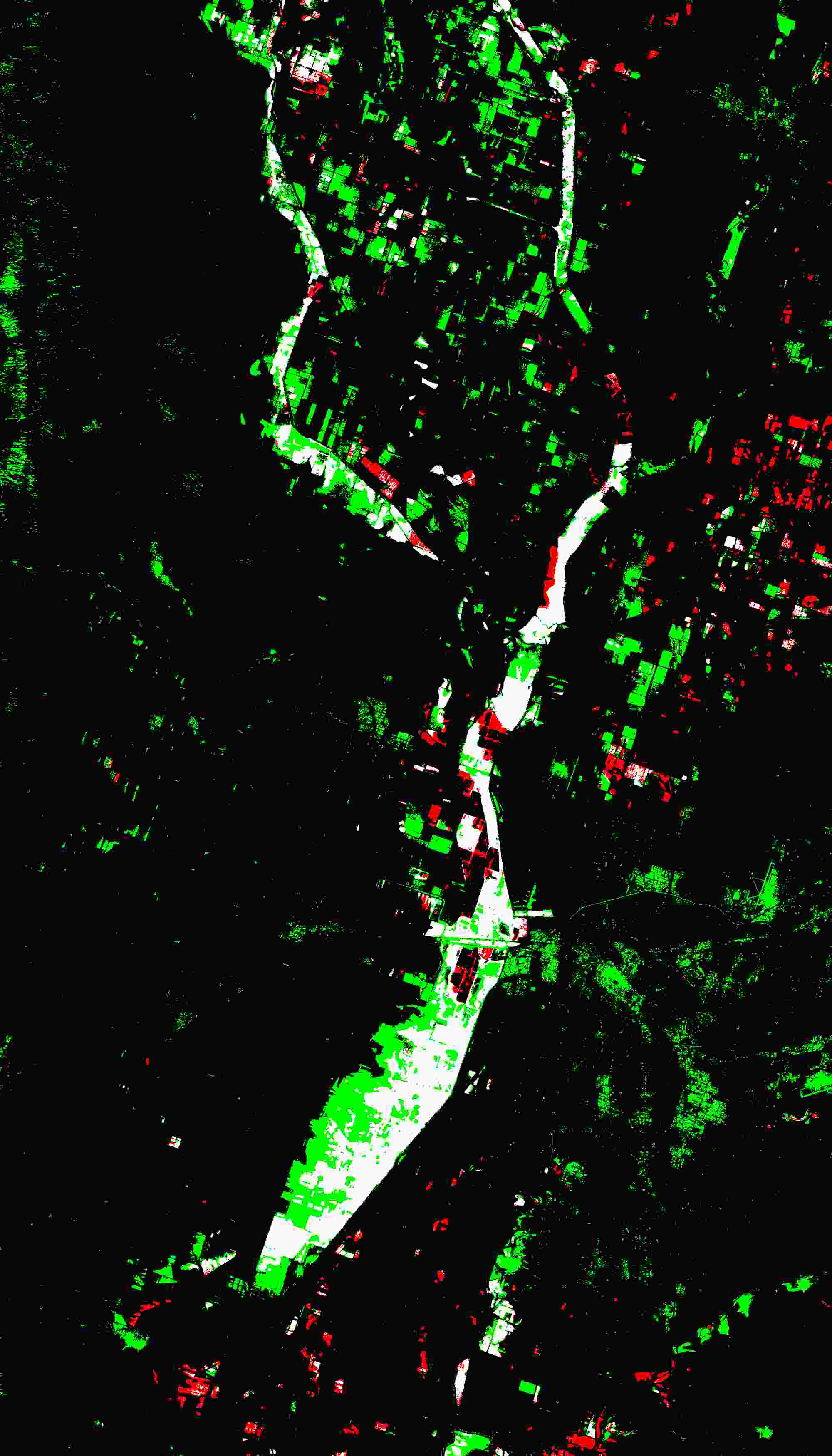}
\caption{\centering Confusion map, RFR, \newline $M=10^5$, $k=20$}
\label{fig:confusion_cal}
\end{subfigure}
\end{center}

\caption{Some examples from Table \ref{tab:Results_cal}. The worst regression in all terms of accuracy: GPR, $M=10^4$, $k=10$ (a); A good example of regression: RFR, $M=10^5$, $k=20$ (b); Its corresponding output after filtering (c); The corresponding confusion map after thresholding (d). White: TP; Green: FP; Red: FN; Black: TN.}
\label{fig:results2}
\end{figure*}

Table \ref{tab:Results_cal} summarises the results on this dataset.
For the same reasons as before, only RFR and HPT were used to perform the regression with $M=10^5$,
given that the computational burden of the other two methods is even larger with this dataset.
Some examples are shown in Fig.\ \ref{fig:results2}.
Generally speaking, every configuration achieved good results in terms of AUC and OA.
Again, the least satisfactory outcomes are the ones related to the GPR.
Although its AUC and OA are acceptable, this method seems to struggle to converge to an optimal solution, if not even a reasonable one, when inspecting its KC value for all the combinations of $k$ and $M$.
Nevertheless, the KC only goes between 0.3 and 0.46 for the rest of the of the regression methods.
This is due to the high amount of false positives, as it can be noticed in Fig.\ \ref{fig:confusion_cal}, which may be connected to the reliability of the ground truth.
Unfortunately, even a few days difference between $\boldsymbol{Y}$ and the second SAR image might imply changes.
For example, we argue that the large patch of false positives right above the flooded area in the south of the image in Fig.\ \ref{fig:confusion_cal} is actually representing real floods in $\boldsymbol{Y}$ (Fig.\ \ref{fig2:S1A}).
Anyway, in this scenario the RFR showed to be slightly better than the HPT and SVR, but at the same time to be significantly faster, especially for $M=10^5$.

\section{Conclusions}\label{sec:concl}

In this paper, we proposed an unsupervised CD framework based on the comparison of affinity matrices and on image regression.
We evaluated and compared the performance obtained with four different regression methods.
Experiments on two datasets proved the effectiveness of the methodology, both for the self-supervised training set selection and the detection of changes across the two heterogeneous images.
The consistency of the results underlines how a good selection of the training set is more crucial than the specific image regression method used afterwards.

On one hand, the results of our experiments both for the self-supervision phase and for the final change map show an improvement as the window size $k$ increases.
However, a larger $k$ increases the computational load and memory requirements.
Moreover, the algorithm we have designed to compare affinity matrices across images is intended to capture the local structure and information rather than the global one.
On the other hand, a smaller window size may not provide any information at all, especially for imagery with higher resolution.
Potentially, a too small patch could simply have no structure, and the corresponding affinity matrix would capture only noise, corrupting the final output.
The optimal $k$ can be found by means of cross-validation or, as an alternative, one can make use of a more robust ensemble approach to combine the outcomes based on different window sizes.

Concerning the image regression, we note that RF regression was able to reach performance comparable to state-of-the-art regression methods, but requires a considerably shorter computation time and the hyperparameters are robust and easy to tune.
On the contrary, the HPT and the SVM regression are generally much slower.
The former has hyperparameters which are quite easy to tune, but it spends a lot of time on the search for $K$-nearest-neighbours during the training phase.
The latter instead requires to carefully select three sensitive hyperparameters through a fine-gridded and costly search.
GP regression requires the computation of all the elements of the kernel matrices, and all the matrix multiplications to obtain the final output for each input data point, so it implicates a large computational burden, especially for large datasets with many channels involved.
In conclusion, we recommend to use RF to perform regression, given their reliability, robustness, speed, and ease of tuning.

A future work would be to explore new approaches to improve the results obtained by Algorithm \ref{alg:aff} and to lower its computational cost.
Besides the advantages for training set selection, such an improvement could potentially turn the affinity-based analysis into a stand-alone method for CD.
Also, a sampling strategy ensuring that the entire feature space of the input is covered by the training set can potentially improve the regression phase.
The number of processed patches may be reduced by shifting the patch multiple pixels instead of one pixel at the time.
This will speed up computations, but the effects on the training set selection must be investigated.





\bibliographystyle{IEEEtran}
\bibliography{references}

\end{document}